\documentclass[sigconf]{acmart}
\usepackage{booktabs} 
\usepackage{amsmath}
\usepackage{algorithmic}
\usepackage{algorithm}
\usepackage{subfig}
\usepackage{multirow}
\usepackage{color}
\usepackage{natbib}
\usepackage{xstring}
\DeclareMathOperator*{\argmax}{\arg\,\max}

\usepackage{xcolor}

\usepackage[T1]{fontenc}
\usepackage{balance}

\editor{Jennifer B. Sartor}
\editor{Theo D'Hondt}
\editor{Wolfgang De Meuter}

\begin{document}
\title{Towards Query Efficient Black-box Attacks: \\
An Input-free Perspective}

 \author{Yali~Du$^{\ast 123}$\ \ Meng~Fang$^{\ast 4}$\ \ Jinfeng~Yi$^5$\ \ Jun~Cheng$^{26}$\ \ Dacheng Tao$^3$}
 \thanks{$\ast$ Contributed equally}
 \affiliation{
 	\institution{
 $^1$ Centre for Artificial Intelligence, FEIT, University of Technology Sydney, Australia\\
  $^2$   Guangdong Key Lab of Robotics and Intelligent System, Shenzhen Institutes of Advanced Technology, CAS, China\\
  $^3$ UBTECH Sydney AI Centre, School of IT, FEIT, the University of Sydney,  Australia \\
 $^4$  Tencent AI Lab, Shenzhen, China \ \  $^5$  JD AI Research, Beijing, China\ \  $^6$ The Chinese University of Hong Kong \\
  yali.du@student.uts.edu.au, mfang@tencent.com,yijinfeng@jd.com, \\jun.cheng@siat.ac.cn, dacheng.tao@sydney.edu.au 
   }
}







\renewcommand{\shortauthors}{Y. Du et al.}

\begin{abstract}
Recent studies have highlighted that deep neural networks (DNNs) are vulnerable to adversarial attacks, even in a black-box scenario. However, most of the existing black-box attack algorithms need to make a huge amount of queries to perform attacks, which is not practical in the real world. We note one of the main reasons for the massive queries is that the adversarial example is required to be visually similar to the original image, but in many cases, how adversarial examples look like does not matter much. It inspires us to introduce a new attack called \emph{input-free} attack, under which an adversary can choose an arbitrary image to start with and is allowed to add perceptible perturbations on it. Following this approach, we propose two techniques to significantly reduce the query complexity. First, we initialize an adversarial example with a gray color image on which every pixel has roughly the same importance for the target model. Then we shrink the dimension of the attack space by perturbing a small region and tiling it to cover the input image. To make our algorithm more effective, we stabilize a projected gradient ascent algorithm with momentum, and also propose a heuristic approach for region size selection. 
Through extensive experiments, we show that with only 1,701 queries on average, we can perturb a gray image to any target class of ImageNet with a 100\% success rate on InceptionV3. Besides, our algorithm has successfully defeated two real-world systems, the Clarifai food detection API and the Baidu Animal Identification API.


\end{abstract}
\copyrightyear{2018} 
\acmYear{2018} 
\setcopyright{acmcopyright}
\acmConference[AISec '18]{11th ACM Workshop on Artificial Intelligence and Security}{October 19, 2018}{Toronto, ON, Canada}
\acmBooktitle{11th ACM Workshop on Artificial Intelligence and Security (AISec '18), October 19, 2018, Toronto, ON, Canada}
\acmPrice{15.00}
\acmDOI{10.1145/3270101.3270106}
\acmISBN{978-1-4503-6004-3/18/10}


\maketitle

\section{Introduction}

Given a legitimate example, adversarial attacks aim to generate adversarial examples by adding human-imperceptible perturbations~\citep{carlini2017towards,moosavi2016deepfool,goodfellow2014explaining}. Depending on how much information an adversary can access to, attacks can be classified as white-box or black-box. When performing a white-box attack, the adversary has full control and complete access to the target model~\citep{carlini2017towards,moosavi2016deepfool,goodfellow2014explaining}. While performing a black-box attack, the adversary has no access to the explicit knowledge of the underlying model, but it can make queries to obtain corresponding outputs~\citep{brendel2017decision,chen2017zoo,ilyas2018black,hayes2017machine,liu2016delving,tu2018autozoom}.


Compared with white-box attacks, black-box attacks are more realistic and more challenging. Although many black-box attack methods have been proposed, most of them suffer from high query complexity which is proportional to the size of the image. For example, attacking a $299\times 299 \times 3$ ImageNet image can usually take hundreds of thousands of queries, or even more. Indeed, most of these approaches would fail by simply setting an upper limit on the number of queries one can make. 
Note that an important reason for the high query complexity is that the adversarial example is required to be visually similar to the original image. This is because each pixel on the original image usually plays a significantly different role on the model's outputs, and the adversary may need to query all of the pixels to understand their roles. However, 
a primary goal of black-box attack is to break down the underlying target model with as little cost as possible. When this goal is achieved, how adversarial examples look like does not matter too much. For this reason, we re-evaluate the necessity of the common practice that adversarial examples should be visually similar to the meaningful input images.

In this work, we focus on a specific type of attack, which we refer to as ``\emph{input-free}" attack. With this attack, adversaries are free to choose any initial image to perform attacks. A typical initial image is a gray color image with each pixel value equal to 128. One significant advantage of this kind of initial image is that all of the pixels have exactly the same intensity, thus they should have the same importance. Note that many black-box adversarial attack algorithms need to make a large number of queries to recognize the importance of each pixel, starting from a gray color image can greatly reduce the query complexity. Besides, ``\emph{input-free}" attacks are more desirable than traditional adversarial attacks when benign examples are not accessible. For instance, an adversary may have no access to any rightful faces that can unlock a face recognition system.

 Under \emph{input-free} settings, the adversarial attack becomes slightly simpler since the distortion constraint is removed. In other words, an \emph{input-free} attack can choose an arbitrary image to start with and is allowed to add perceptible perturbations on it. The effectiveness of an \emph{input-free} attack is measured by the cost spent during the attack. The cost is usually represented by the number of queries, and the lower the number is, the more effective the attack is. 
We note that \citep{nguyen2015deep} also lies in the scope of \emph{input-free} attack, in which the adversarial examples were generated by starting from a random image. However, \citep{nguyen2015deep} doesn't focus on improving the query complexity of black-box attacks, thus is different from our concern.   
{Recently, \citet{ilyas2018black} proposed to adopt Natural Evolutionary Strategies (NES) \citep{wierstra2014natural} to optimize black-box attacks. It shows that fewer queries are needed for the gradient estimation than the finite-difference methods \citep{chen2017zoo}. 
Based on the NES approach \citep{ilyas2018black,wierstra2014natural}, we propose two techniques to further improve the query complexity, namely gray image attack and region-based attack, which are explained as follows.}

\paragraph{Gray image attack}
First, we choose a gray color image with uniform pixel values as the starting point, and then keep adding noises on it until it is successfully classified into the target class. The gray image does not contain any human perceivable information, which is different from the traditional perturbation constraint on the adversarial examples. {In other words, this kind of attack abandons the similarity constraint that most standard attacks used~\citep{chen2017zoo,ilyas2018black}.} 
Since all of the pixels are initialized to be the same value, they would be almost equally important for the network output. In this case, a gray image evades the strength of convolution neural networks in extracting representational features from input images, rendering it easier to be classified into other classes. Note that the gray color image is not the only choice for \textit{input-free} attack. We have tried other natural images such as rocks and woods and they all lead to satisfactory performance.

\paragraph{Region-based attack}
The equivalence of pixels leads us to consider whether it is necessary to query the whole image to synthesize an adversarial example. When developing an image recognition system, the model's sensitivity to common object deformations such as translation and scaling are implicitly considered in general. 
Inspired by this fact, we do not need to query the whole image. Instead, we perturb a much smaller region and then tile it to cover the whole image to match the input dimension. This operation can significantly reduce the dimension of attack space, and thus improve the query complexity. 

\begin{figure*}[th!]
	\centering
	\includegraphics[width=160mm,]{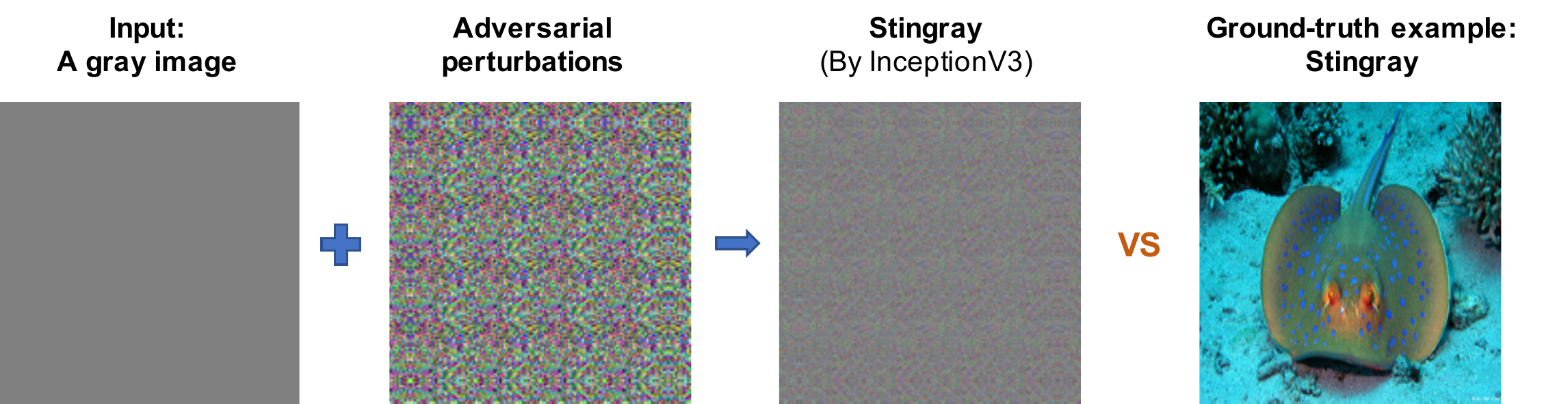}
	\caption{Illustration of adversarial examples on ImageNet for \emph{Stringray} class with $735$ queries. The perturbations show repeated patterns of noises and the resulting image contains the texture and edge informaiton of  \emph{Stringray} class.  }\label{fig1:res}
\end{figure*}

We name the proposed attack method as \emph{Region Attack}, and evaluate it on InceptionV3 \citep{szegedy2016rethinking}, a popular classifier trained on ImageNet \citep{deng2009imagenet}. Figure \ref{fig1:res} shows one example of the proposed attack algorithm. The adversarial example is initialized by a gray image, and after only $735$ queries, the output image is misclassified to the target class \emph{Stingray}. Our empirical study shows that on average, the proposed attack algorithm only needs 1,701 queries to reach 100\% success rate with target attack on ImageNet.   
Furthermore, We test the proposed Region Attack on some challenging real-world systems, including Clarifai food detection API\footnote{https://clarifai.com/models} and Baidu Animal Identification API\footnote{https://cloud.baidu.com/product/imagerecognition/fine$_{-}$grained}. These two APIs only return a top list of labels and scores that do not sum to one. We show that Region Attack can still succeed within a small amount of queries. 

Our main contributions are summarized as follows:
\begin{itemize}
	\item We show that the query complexity of black-box attacks can be significantly improved by starting from a gray image. This opens up a prominent direction of performing efficient attacks by designing proper input images. 
	\item 
	{We shrink the dimension of the attack space by perturbing a region and tiling it to the whole image. Compared with the baseline models, this technique reduces average query requirement by at least a factor of two. Besides, we also propose a heuristic algorithm for region size selection.}
	\item {We show that momentum-based stochastic gradient descent (SGD) can stabilize the update of the search distribution, which was also examined to boost traditional gradient-based adversarial attacks \citep{dong2018boosting,carlini2017towards}. }
    \item We show that the proposed model could successfully attack commercial online classification systems.
   To the best of our knowledge, the proposed attack model is one of the pioneering works that reduces query complexity from the perspective of \textit{input-free} attack. 
\end{itemize}

\section{Related Works}\label{sec:related-works}
The research of adversarial attacks is rooted in the need for understanding the security of machine learning models and hence deploying reliable systems in real-world applications \citep{barreno2006can,barreno2010security}. After  \citet{szegedy2014intriguing} pointed out the existence of adversarial examples, there is a long line of works  \citep{goodfellow2014explaining,moosavi2016deepfool,brendel2017decision,papernot2016limitations,papernot2017practical,ilyas2018black,chen2017zoo,carlini2017towards,engstrom2017rotation} that produce adversarial examples in different settings.

The adversarial attack approaches fall into two categories: white-box and black-box attacks. White-box attacks \citep{szegedy2014intriguing,goodfellow2014explaining,moosavi2016deepfool,carlini2017towards,papernot2016limitations} explore the target model's architecture and parameter weights to find the gradient direction to update adversarial perturbations. However, when target models do not release their architectures to the public, it is impossible for white-box attacks to back propagate the gradients of the network output with regard to the input image. 

Black-box attacks do not require the architecture or parameters of the target model, but need to query the target model to obtain feedback on the adversarial examples.
The transferability of adversarial examples \citep{szegedy2014intriguing}, \citep{papernot2017practical} can construct a substitute model to perform white-box attacks and take the query results from the target model to refine the substitute model's classification boundary. Since training a substitute model is cumbersome and adversarial examples do not always transfer well, black-box attacks based solely on queries are in highly demand.

When black-box attacks have only query access to a target model \citep{hayes2017machine,liu2016delving,ilyas2018black,chen2017zoo}, to imitate gradient information, \citep{chen2017zoo,ilyas2018black} performed black-box attacks by zeroth order optimization. 
\citep{chen2017zoo} proposed to estimate gradients via coordinate-wise finite differences based on probability vector of different classes. 
\citep{ilyas2018black} maximized the expected value of loss function by a search distribution, and approximated gradients by Natural Evolution Strategies \citep{wierstra2014natural} through a Gaussian search distribution. These methods rely on probabilities show the relative goodness of adversarial examples. \citep{brendel2017decision} does not require probability to estimate gradients but only labels of the target model. It initialized the adversarial example by an image in target class and greedily reduced the distance between the adversarial example and the image from the target class, and preserved the prediction results of the adversarial example in the target class. The number of queries of this algorithm was as high as 400,000 for falsifying an adversarial example for a simple neural classifier on CIFAR10.

Existing research under \textit{input-free} attack is scarce. \citet{nguyen2015deep} evolved from uniform random noise and compressible patterns to fool deep neural networks. \citep{nguyen2015deep} aimed to produce adversarial examples with high confidence, but the efficiency of the algorithm was not guaranteed.   
\citet{ilyas2018black} claimed that maximizing the expected value of fitness function by Natural Evolution Strategies requires fewer queries than typical finite-difference methods. However, it still needs a median of more than 10,000 queries to attack InceptionV3 classifier on ImageNet. \citep{engstrom2017rotation} studied rotations and translations for adversarial attacks which supposedly requires fewer queries, but the success of its attacks was not guaranteed.

{The proposed region-based attack restricts the dimension of the attack space; in this respect,  \citep{chen2017zoo} adopted an attack-space dimension technique which first queries a small-sized image to generate adversarial noises and then upscales it to the dimension of the original image. \cite{su2017one} performed attacks by modifying one or few pixels but the success rate of targeted attacks was low.
}

\section{Query Efficient Black-Box Attacks based on Natural Evolution Strategies}\label{sec:query-efficient-attack}
\subsection{Preliminaries}
\paragraph{Deep Neural Networks and Adversarial Attacks}
Following the naming conventions in \citep{carlini2017towards}, we denote  $x\in[0,1]^{d}$ as an input image,  $F(x)$ as the output of the softmax layer and $Z(x)$ as the logit layer which is the output of the last layer before softmax. $C(x)$ gives final label prediction. Specifically:
\begin{align*}
	F(x)&=\text{softmax}(Z(x)),\\
    C(x) &= \argmax _{i}[F(x)]_i.
\end{align*}
Assume original prediction on $x$ satisfies that $C(x)=y$, an adversary aims to either reduce the classifier's confidence on correct labels or alter the classifier's prediction to any specified class by adding spuriously crafted noise on the image $x$. 
In untargeted attacks, an adversary intends to perturb the input $x$ with maliciously crafted noises $\delta$ such that 
\begin{equation*}
	C(x+\delta)\neq y \ \ \text{and} \ \ \|\delta\|<c,
\end{equation*}
where $\|\cdot\|$ is a norm chosen by the attacker, such as $L_2, L_{\infty}$ and $c$ is a pre-specified constant. 
Targeted attacks craft adversarial perturbations that explicitly classify an input image into another specified class $y'$, which is formulated as follows:
\begin{equation}\label{eq:attack-original}
	C(x+\delta)= y', y'\neq y,  \ \ \text{and} \ \ \|\delta\|<c.
\end{equation}
The formulation in Eq. (\ref{eq:attack-original}) can not be directly solved by existing algorithms, as the equality constraint $C(x+\delta)=y'$ is highly non-linear. Accordingly, one can express it in an equivalent form that is more suitable for optimization \citep{carlini2017towards}:
\begin{equation}\label{eq:attack-opt-easy}
	f(x, y')=\max \{ \max_{t\neq y'} \log[F(x)]_{t} - \log[F(x)]_{y'}, - \kappa  \},
\end{equation}
where $\kappa$ is a suitable constant.
Eq. (\ref{eq:attack-opt-easy}) is more suitable for optimization and has been applied to both white-box and black-box attacks. When performing white-box attacks, one can directly take gradient derivation to image $x$ to minimize the loss function $f(x, y')$. When performing black-box attacks, gradient can be approximated by pixel-wise finite differences. 

\paragraph{Black-box optimization by Natural Evolution Strategies}
To perform a black-box attack, one needs to search the input space for adversarial noise which satisfies Eq. (\ref{eq:attack-original}).  Existing works mainly solve Eq. (\ref{eq:attack-opt-easy}) through minimizing the objective function. { In this work we adopt Natural Evolutionary Strategies \citep{wierstra2014natural} which is used in \citep{ilyas2018black} to estimate gradients from probabilities of predictions for derivative-free optimization.}
Instead of maximizing the objective function directly, NES optimize the expected value of a fitness function under a search distribution.

The procedure of optimization by NES can be summarized as four steps from \citep{wierstra2014natural}: (i) one samples a batch of search points from the chosen parameterized search distribution and evaluates the fitness function on them; (ii) NES estimates search gradients for the parameters of the search distribution; (iii) NES performs gradient ascent step along the search gradient direction; (iv) NES repeats the above three steps until a stopping criterion is satisfied. 
Denote the fitness function as $J(\theta)=P(y|\theta)$ and $x$ as parameters for search distribution, the formulation of search gradients can be derived as follows, per \citep{ilyas2018black,wierstra2014natural}: 
\begin{align*}
	\mathbb{E}_{\pi(\theta|x)}[J(\theta)]&=\int J(\theta) \pi(\theta|x)d\theta \nonumber\\
	\nabla_{x} \mathbb{E}_{\pi(\theta|x)}[J(\theta)] & = \nabla_{x}\int J(\theta)\pi(\theta|x)d\theta \nonumber\\
	&=\int J(\theta)\nabla_{x}\pi(\theta|x)d\theta.
\end{align*}
Applying the ``log-likelihood'' trick yields:
\begin{align}\label{eq:nes-grad}
	\nabla_{x} \mathbb{E}_{\pi(\theta|x)}[J(\theta)] & = \int J(\theta)\frac{\pi(\theta|x)}{\pi(\theta|x)}\nabla_{x}\pi(\theta|x)d\theta \nonumber\\
	&=\int \pi(\theta|x)J(\theta)\nabla_{x}\log\pi(\theta|x)d\theta \nonumber\\
	&=\mathbb{E}_{\pi(\theta|x)}[J(\theta)\nabla_{x}\log\pi(\theta|x)].
\end{align}
The derivations in Eq. (\ref{eq:nes-grad}) gives mathematical foundations for steps (ii) and (iii). From samples $\theta_1, ..., \theta_n$, the search gradient for the loss function $J(\theta)$ is reduced to this form:
\begin{equation}
	\nabla_{x}\mathbb{E} [J(\theta)] \approx \frac{1}{ n} \sum_{k=1}^{n}J(\theta_k)\nabla_{\theta}\log\pi(\theta_k|x).
\end{equation}

To avoid non-smoothness of the fitness function, \emph{local reparameterization trick} \citep{salimans2017evolution,ilyas2018black} can be used by setting $\theta = x +\sigma \delta$ with $\delta\sim \mathcal{N}(0, I)$. In this case, sampling $\theta$ is equal to sampling $\delta$, and at each step, we are literally searching the neighborhood of $x$ for a better solution. Denote $n$ as batch size which is the number of search points used in each step for estimating gradients, $\nabla_{x}\mathbb{E} [J(\theta)]$ is approximated by:
\begin{equation}
	\nabla_{x}\mathbb{E} [J(\theta)]\approx \frac{1}{\sigma n}\sum_{k=1}^{n}J(\theta + \sigma \delta_k)\cdot \delta_k.
\end{equation}
Rather than sampling $n$ search points independently,  we follow \citep{wierstra2014natural,salimans2017evolution,ilyas2018black} and implement an antithetic sampling strategy such that $\delta_k \sim \mathcal{N}(0, I), \text{for } k =1, 2, ..., \frac{n}{2}$, and $\delta_k = -\delta_{n-k+1}, \text{for } k =\frac{n}{2}+1, ..., n$. The search gradient under Gaussian distribution can now be estimated by:
\begin{equation}\label{eq:nes-grad-approx}
	\nabla_{x}\mathbb{E}[J(\theta)] \approx \frac{1}{\sigma n} \sum_{k=1}^{n/2}(\delta_k J(\theta+\sigma\delta_k)-\delta_k J(\theta-\sigma\delta_k)).
\end{equation}

\begin{figure*}[th!b]
	\centering
	\includegraphics[width=160mm,]{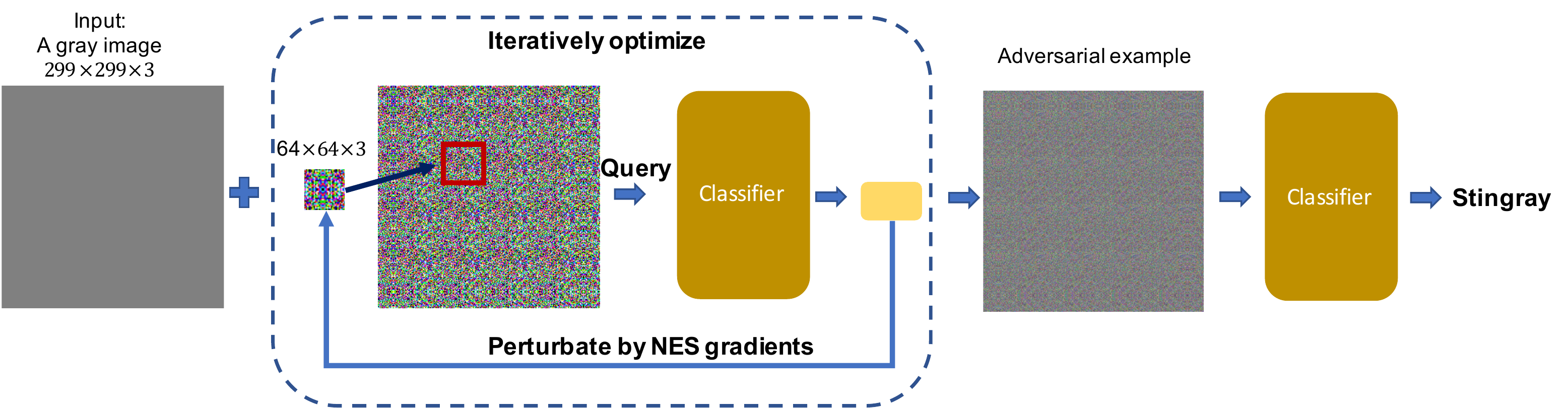}
	\caption{Overview of the proposed Region Attack algorithm. The input is a gray image. By querying the black-box classifier, we obtain the gradients to update a region on the image in the red box. This process is repeated until we reach an adversarial example that is identified to be the target class.
	}\label{fig0:sys-graph}
\end{figure*}

\subsection{The Proposed Attacks by Gray Images}
With the proposed black-box attack, the adversary does not have access to any natural input, which is different from the existing attack algorithms. By normalizing the pixel values within $[0,1]$, we initialize the input to the target model with $x=[x_{ij}]_{n\times m}, x_{ij}=0.5 \ \  \forall (i,j)$ and call it ``gray image''. This initialization leaves us much flexibility since we can perturb the pixel values to be either brighter (closer to 1) or darker (closer to 0) to falsify an example into target class. One can choose different methods to initialize the input other than a gray image.

We first define a fitness function which satisfies that its maximum is obtained when $C(x+\delta)=y'$ and it is dependent on the input to the target  model. Denote input $\theta=x+\delta$, the fitness function is defined as:
\begin{equation}
	J(\theta)= P(y'|\theta)\approx [F(\theta)]_{y'}.
\end{equation}

{We adopt NES optimization and maximize $P(y'|\theta)$ directly until $C(x+\delta)= y'$. 
We choose Gaussian as search distribution for $\theta$ which provides a simple form for gradient estimation \citep{ilyas2018black,salimans2017evolution}.} For a Gaussian distribution $\theta\sim \mathcal{N}(\psi_d, \sigma_{d\times d})$, parameters to be optimized are the mean $ \psi_d$  and covariance $\sigma_{d\times d}$. By performing gradient ascent for parameters, the expected value of fitness function is maximized.

Aside from Gaussian, other distributions whose derivatives of its log-density with regard to their parameters are attainable can be used as search distributions, such as the Laplace distribution. By setting $\theta = x+b\delta$ with $\delta \sim \text{Laplace}(0, 1)$,  the search gradient of the fitness function can be estimated in the following form:
\begin{equation}
	\nabla_{x}\mathbb{E} [J(\theta)]\approx \frac{1}{b n}\sum_{k=1}^{n}J(\theta + b \delta_k)\cdot \text{sign}(\delta_k).
\end{equation}

Note that we adopt a region-based attack algorithm, so gradients are dependent on 
regions. We denote the gradient as $g(h, w)$ with $h, w$ denoting region size. The details will be expanded in Section \ref{sec:region-based-attack}.
After evaluating the gradients, we perform projected gradient ascent as in \citep{madry2017towards,ilyas2018black} by using the sign of the obtained gradients to update $x$:
\begin{equation}\label{eq:graddescent}
	x^{t}=\Pi_{[0, 1]} (x^{t-1}+\gamma\text{sign}(g_t(h,w)),
\end{equation}
where $\gamma$ denotes the learning rate and $\Pi_{[0,1]}$ denotes the projection operator that clips pixel values into $[0,1]$. 
The adversarial image is updated by the search gradients until the adversarial goal is achieved. 

We show the overall process of Region Attack in Fig. \ref{fig0:sys-graph}. The adversarial example is initialized with a gray image. 
We produce search gradients in a small region indicated by the red box and tile it to the whole image; thus, the gradient information for each region in the image is exactly the same. After black-box optimization, we obtain the adversarial example with clear texture and edging information showing in each region.


{Several issues require our attention for implementing Region Attacks. On one hand, how to set up an appropriate region size? On the other hand, since gradients are estimated based on the finite sampling of search points, the variance will be introduced and prohibits convergence of the algorithm. How to tame this challenge? }
First, we summarize the general procedures for the gray image attack in Algorithm \ref{alg:all}, and then discuss the key steps in the remaining section. Step 1, 3, and 4 in Algorithm \ref{alg:all} will be further explained in Section \ref{sec:region-based-attack}, \ref{sec:region-size-selection}, and \ref{sec:momentum-based gradient-descent}.

\begin{algorithm}[!htb]
	\caption{\label{alg:all} Canonical Procedures for Query Efficient Attack}
	\begin{algorithmic}[1]
		\REQUIRE target class $y'$, gray image $x$, max iteration $maxiter$
		\ENSURE adversarial image $x'$ such at $C(x') = y'$ 
		\STATE set up the region size $(h,w)$
		\STATE \textbf{for } $t= 1,2,...,maxiter$ \textbf{do} 
		\STATE estimate gradient $g_t(h, w)$ by region mutation
		\STATE     $x^{t}=\Pi_{[0, 1]} (x^{t-1}+\gamma\text{sign}(g_t(h,w)))$
		\STATE \textbf{if} $C(x^t )= y':$
		\\ \ \ \  success = True
		\\ \ \ \  return $x^{i}$
		\STATE \textbf{end if}
		\STATE \textbf{end for}
	\end{algorithmic}
\end{algorithm}

\subsection{Gradient Estimation based on Region Mutation}\label{sec:region-based-attack}
We denote a perturbed region as $R=(p_1, p_2, h, w)$, where $(p_1, p_2)$ is the position of the pixel at the top left corner of the chosen region, and $h, w$ indicate the height and width of the perturbed region. We tile the region to the whole image such that every region of size $h
\times w$ has the same pixel values. Now the dimension of the attack space we need to optimize is $h \times w$; in fact, adopting a small region size will lead to a reduction on dimension of the attack space.  If $h\times w$ cannot be divided by $d^2$, we place $\lfloor \frac{d^2}{hw}\rfloor$ regions in the middle of the image and adopt symmetric padding to complete the whole image.

Based on the above discussion, the gradient estimation over the whole image is aggregated by gradients on different regions. We summarize the gradient estimation procedures in Algorithm \ref{alg:grad} based on antithetic sampling \citep{ilyas2018black,salimans2017evolution}, which will be used to replace Step 3 in Algorithm \ref{alg:all}.

\begin{algorithm}[th!b]
	\caption{\label{alg:grad} NES Gradient  Estimation for Region Mutation}
	\begin{algorithmic}[1]
		\REQUIRE target class $t$, image $x$ of dimension $d$, classifier $P(y|x)$, batch size $n$, region size $h, w$
		\ENSURE gradient estimation $g$
		\STATE $g\leftarrow\mathbf{0}_{d}$
		\STATE \textbf{for } $i= 1,2,...,\frac{n}{2}$ \textbf{do} 
		\STATE sample $\epsilon_k\sim N(0_{hw}, \delta_{hw\times hw})$
		\STATE repeat $\epsilon_k$ by $\lceil \frac{d^2}{hw} \rceil$ times thus obtain $\tilde{\epsilon}_k\in \mathbb{R}^{d}$
		\STATE $g = g+\frac{1}{\sigma}P(y|x+\sigma\cdot \tilde{\epsilon}_k)\cdot \tilde{\epsilon}_k$
		\STATE $g = g-\frac{1}{\sigma} P(y|x-\sigma\cdot \tilde{\epsilon}_k)\cdot \tilde{\epsilon}_k$
		\STATE \textbf{end for}
		\STATE \textbf{ return} $\frac{g}{n}$
	\end{algorithmic}
\end{algorithm}

\subsection{Active Selection on Region Sizes}\label{sec:region-size-selection}
We propose a heuristic approach based on a limited query budget. We use the variation of entropy to measure whether a chosen size is effective for Region Attack on the target model. We set the batch size as $n_0$, and take $t_0$ iterations in this warm-up section. Define the entropy of the target model's prediction under current region size as:
\begin{equation}
	h(F(x)) = \sum_{i=1}^{n}[F(x)]_i\log [F(x)]_i.
\end{equation}
We use its variation $\Delta h$ to measure the effectiveness of current selected size, which is defined as 
\begin{equation}
	\Delta h = h_{t_0}-\min(h_j), \ j \in \{1,..., t_0\},
\end{equation}
where $h_{j}$ denote the entropy calculated after $j-$th updating step.

The larger the variation of entropy is, the more capable we are to alter the prediction of the original input. This ensures that we are more likely to reach the target goal under the same number of queries. In fact, most of the targeted attacks would experience such two stages that the entropy $h_j$ of predicted probability vector first goes down and then goes up. 
{In the first stage, Region Attack reduces the labeling information of the original prediction on the gray image, which leads to the decrease of $h_j$. In the second stage, Region Attack raises the labeling information of the target class which corresponds to the increase of $h_j$.
The steps of  this approach are listed as follows:
\begin{enumerate}
    \item we perform momentum gradient ascent for $t_0$ iterations and record the output of softmax layer $F(x)$;
    \item we  calculate the entropy of $F(x)$ after each iteration to obtain $h_j$;
    \item we estimate the variation of entropies of these $t_0$ iterations and return the size that corresponds to the largest variance.
\end{enumerate}
The region size used in Step 1 in Algorithm \ref{alg:all} can be set up by these procedures.

\subsection{Momentum-based Gradient Ascent}\label{sec:momentum-based gradient-descent}
Since one can not search the whole distribution to calculate the accurate gradient, one always samples a batch of points from search distribution to estimate gradients.  Due to the limited sample size for one batch, the variance would be introduced into the gradients and hurts the convergence of the algorithm. Increasing the sample size would reduce the variance for each estimation of the gradient. However, monotonically increasing the sample size for one draw will increase query complexity. {Note that the momentum-based gradient descent has been verified in \citep{dong2018boosting,carlini2017towards} to be effective in stabilizing update directions of objective function which is mentioned but not discussed in \citep{ilyas2018black}. Following this route, we use the approach of momentum-based gradient ascent to update the mean of the search distribution, which is exactly the adversarial image:}
\begin{equation}\label{eq:mom:graddescent}
    \begin{aligned}
        &    g_{t}=\eta g_{t-1} + (1-\eta)\nabla_{\theta}J(\theta)\\
        &    x^{t}=\Pi_{[0, 1]} (x^{t-1}+\gamma\text{sign}(g_t)).
    \end{aligned}
\end{equation}
One thing to mention is that we do not follow the convention to set $\eta$ to 0.9 or a similar value, we instead set it to be around $0.5$. Since we use the sign of gradients to update adversarial perturbations, setting $\eta=0.9$ will lead to the situation that the algorithm keeps exploiting in the same direction without exploring new directions.
	
\section{Partial-information attack}\label{sec:partial-info}
In addition to the discussion in Section \ref{sec:query-efficient-attack}, we note that many open cloud APIs provide only TOP-$k$  scores of detected labels that are not summed to 1. This situation is difficult since neither the number of labels nor the probability of all labels is given. Furthermore, in our procedure, an attacker does not have access to benign inputs, and a random initialization might lead to the prediction as \emph{``non-object''} or \emph{``background''}. This makes it more challenging than the partial-information attack discussed in \citep{ilyas2018black}.   
	
We deploy a two stage process for efficient attacks. Firstly, we treat a target model as a binary classifier which produces the label of either \emph{``non-object''} or  \emph{``is object''}. Now it is possible to perform an untargeted attack by minimizing the probability of $P(\textit{non-object}|x)$. We minimize  $P(\textit{non-object}|x)$ until it falls out of TOP-K predictions. Secondly, we choose a legal label from TOP-K\{$P(\cdot|x)$\} as a target class $y'$ and maximize $P(y'|x)$ until the target model confidently recognizes $x'$ as an legal object. In the first stage, we reverse gradient ascent steps in Section \ref{sec:region-based-attack} to gradient descent to minimize $P(\textit{non-object}|x)$. {In the second stage, we follow the two-step optimization procedure used in \citep{ilyas2018black,brendel2017decision} which alternates between maximizing probability of $y'$ as $P(y'|x)$ and retaining $y'$ in the TOP-K\{$P(\cdot|x)$\} predictions by shrinking the learning rate. }
	The general procedures are summarized into the following four steps.
	
\fbox{
  \parbox{0.9\columnwidth}{
  \begin{enumerate}
      \item initialize an input $x_0$ and its label $y_0=C(x_0)$ and specify a classifier $P(\cdot|x)$ that returns TOP-K labels and scores;
      \item minimize $P(y_0|x)$ until $y_0\notin$TOP-K\{$P(\cdot|x)$\};
      \item  choose $y'\in$  TOP-K\{$P(\cdot|x)$\} as the target class;
      \item alternate between the following two steps: \\
      (a) maximizing the score of the target class: $x^t= \arg\max_{x}P(y'|x)$;\\
      (b) adjusting the learning rate $\gamma$ such that: $\gamma=\max \gamma' \text{\ s.t. rank}(y'|x^t)\leq k $;
      \item  quit when $P(y'|x)>c$ where $c$ is a specified constant.
  \end{enumerate}
}}
	
\section{Experiments}\label{sec:experiment}
	We examine the proposed algorithm's efficiency in generating target adversarial examples and reducing query complexity. We would like to show that the proposed algorithm achieves a high success rate and leads to a significant reduction in query complexity. Furthermore, we demonstrate that Region Attack can fool many open Cloud APIs with affordable number of queries.
	
\subsection{Baselines}
We adopt two recent attacks ZOO \citep{chen2017zoo} and Query-Limited (QL) \citep{ilyas2018black} as baselines for comparison. 
	{Since the similarity between images is not the main issue in our settings,
	to be fair, by removing the similarity constrain in objective functions,} we construct two stronger baselines ZOO-$L_2$ and QL-$L_{\infty}$ with necessary modifications based on ZOO and QL where ``-'' should be read as ``minus''. 
\begin{itemize}
    \item \textbf{ZOO} \citep{chen2017zoo} estimates gradients by pixel-by-pixel finite differences and adopts Adam optimizer to optimize the adversarial perturbations. It includes $L_2$ norm as distance loss to constrain adversarial examples to not be far away from the initialization. We use its public implementation.
    \item \textbf{ZOO-$L_2$} is constructed by slightly modifying ZOO. Since the proposed setting does not consider distance loss, to make fair comparisons, we remove the distance loss in ZOO and construct a more effective baseline denoted as ZOO-$L_2$.
    \item \textbf{QL} \citep{ilyas2018black} applies Natural Evolutionary Strategies to perform black-box optimization and adopts $L_{\infty}$ to constrain the distance between  adversarial and benign inputs. It reports 1$\sim$2 orders of magnitude reduction on query complexity compared to ZOO. We use its implementation without modifications.
    \item \textbf{QL-$L_{\infty}$} is a slightly modified version of QL with $L_{\infty}$ distance removed, which is supposed to be more query-efficient than QL.  
	\end{itemize} 
	To this end, we have four strong baselines. Since both ZOO and QL implemented attacks on InceptionV3, we will report attack results on this classifier.

\subsection{Setup}

We choose InceptionV3 as the target model with default input size $299\times 299\times 3$, which is consistent with the settings in ZOO and QL. InceptionV3 \citep{szegedy2016rethinking,abadi2016tensorflow} achieves 78\% top-1 accuracy for the ImageNet dataset \citep{russakovsky2015imagenet}. All attacks are restricted to have access only to  the probability vectors of all possible classes. 
The default search distribution used through  experiments is the Gaussian distribution. 

	We produce a gray color image with size $299\times 299\times 3$ and pixel values equal to 128, which are normalized to 0.5 for optimization as the initialized input for all baselines in different targeted attacks. For this same gray image $x$, we choose different target class $t_i, i=1,...,1000$, and construct the test set $\mathcal{S}=\{(x,t_1), ..., (x, t_{1000})\}$. Note that in our experiments, we always use a square region by setting $w=h$ for the object. We will only specify the height of the region in the following.
	
\paragraph{Metrics}
Denote $\Omega$ as the set of successfully attacked target classes and $q_i$ as the query required for attacking $i$-th target class, we define two measurements: success rate $r$ and average number of queries $q_{avg}$:
\begin{align*}
    r&=\frac{\|\Omega\|}{1000}, \\ 
    q_{avg}  &=\frac{\sum_{i\in \Omega}q_i}{\|\Omega\|}.
\end{align*}
Success rate indicates the capability of an algorithm to cheat a classifier. Average number of queries indicates the cost of an algorithm to attack. The fewer the number of queries are, the more efficient an algorithm is. 

Since ZOO and QL punish $L_2$ and $L_{\infty}$ distortion of adversarial examples, we report the averaged $L_2$ and $L_{\infty}$ distortions in Section \ref{sec:exp:full-info}. Denote the $L_2$ and $L_{\infty}$ distortions for $i$-th attack as  $\delta_{i,2}, \delta_{i,\infty}$, the averaged $L_2, L_{\infty}$ distortion are defined as:    
\begin{align*}
\delta_{avg,2}&=\frac{\sum_{i\in \Omega}\delta_{i,2}}{\|\Omega\|}, \\ 
\delta_{avg,\infty}&=\frac{\sum_{i\in \Omega}\delta_{i,\infty}}{\|\Omega\|}.
\end{align*}
Average distortion means the average distance between initial gray image and adversarial examples when an attack succeeds. {Averaged distortions are reported for readers' reference and are not considered to be the main measurement since we consider \textit{input-free} attacks. The implementation code is publicly available.\footnote{https://github.com/yalidu/input-free-attack}}

	\begin{table*}[th!]
		\centering
		\caption{Comparisons between different attacks on success rate, average queries, $L_2$ and $L_{\infty}$ distortions against InceptionV3. $\eta$ stands for momentum factors for momentum based SGD. Line 5 shows Region Attack results based on standard SGD.}
		\label{table:1}
			\begin{tabular}{l|ccccc}
				\hline
				& success rate & \ avg. queries & avg. $L_2$ & avg. $L_{\infty}$ & size selection cost\\ \hline
				ZOO & 99.7\% & 111,218 & 1.666 & 0.0223 & -\\ \hline
				ZOO-$L_2$ & 99.6\% & 77,513 & 2.035 & 0.0238 & -\\ \hline
				QL& 100\% & 3,948 & 14.357 & 0.050 &- \\ \hline
				QL-$L_{\infty}$& 100\% & 3,770 & 18.908 & 0.169 &-\\ \hline
				{\textbf{Region Attack}}  &    \multirow{2}{*}{  \textbf{100\% }}&     \multirow{2}{*}{\textbf{1,959}} &    \multirow{2}{*}{ \textbf{18.222}} & \multirow{2}{*}{\textbf{0.169}} & \multirow{2}{*}{-}\\ 
				($h=64 $ with SGD  ) & & & & &\\\hline
				{\textbf{Region Attack}}  &    \multirow{2}{*}{  \textbf{100\% }}&     \multirow{2}{*}{\textbf{1,701}} &    \multirow{2}{*}{ \textbf{24.323}} &\multirow{2}{*}{\textbf{0.167}} &\multirow{2}{*}{-} \\ 
				($h=64; \eta=0.4$) & & & &\\\hline
				{\textbf{Region Attack}}  &    \multirow{2}{*}{ \textbf{100\% }}&     \multirow{2}{*}{\textbf{1,734}} &    \multirow{2}{*}{ \textbf{35.320}} & \multirow{2}{*}{\textbf{0.208}} &\multirow{2}{*}{-}\\ 
				($h=64; \eta=0.7$) & & & &\\\hline
				{\textbf{Region Attack}}  &    \multirow{2}{*}{  \textbf{100\% }}&     \multirow{2}{*}{\textbf{2,941}} &    \multirow{2}{*}{ \textbf{32.605}} &\multirow{2}{*}{\textbf{0.180}} &\multirow{2}{*}{-}\\ 
				($h=32; \eta=0.7$) & & & &\\\hline
				{\textbf{Region Attack}}  &    \multirow{2}{*}{  \textbf{100\% }}&     \multirow{2}{*}{\textbf{1,945}} &    \multirow{2}{*}{ \textbf{31.976}} & \multirow{2}{*}{\textbf{0.173}} &\multirow{2}{*}{200}\\ 
				{(size selection$; \eta=0.4$)} & & & &\\\hline
				{\textbf{Region Attack}}  &    \multirow{2}{*}{  \textbf{100\% }}&     \multirow{2}{*}{\textbf{1909}} &    \multirow{2}{*}{\textbf{37.590}} & \multirow{2}{*}{\textbf{0.211}} &\multirow{2}{*}{200}\\ 
				(size selection;$ \eta=0.7$) & & & &\\\hline
			\end{tabular}%
	\end{table*}

\subsection{Results on Query Efficient Attack}\label{sec:exp:full-info}
For Region Attack, we choose different region sizes and momentum factors, then we run the proposed threat model and four baselines on the test set $\mathcal{S}$. It is important to note that the algorithms and Region Attack are tested under the same gray image and target class set. The results are reported in Table \ref{table:1}. Some adversarial examples, adversarial noises, and ground-truth images will be shown in the appendix.
The attack is not restricted to gray images. In fact, we tried other natural images like rocks and woods and also obtained good performance. Since texture patterns are not explicitly shown in these perturbed natural images, they are not displayed in the paper.

	In Table \ref{table:1}, the proposed algorithm obtains 100\% success rate which is the same as QL and QL-$L_{\infty}$, but higher than ZOO and ZOO-$L_{2}$. It is shown that for the proposed algorithm with the fixed region size $w=h=64$,  the query complexity is only around 1700 on average, which is 60 times more efficient than ZOO and ZOO-$L_{2}$, and 2 times faster than QL and QL-$L_{\infty}$.
	
	To evaluate distance loss's influence on query complexity, we compare ZOO with ZOO-$L_{2}$ and QL with QL-$L_{\infty}$.
	We observe that without $L_2$ distance loss between benign and adversarial images,  ZOO-$L_2$ requires fewer queries and achieves comparable success rate compared to ZOO. Removing $L_{\infty}$ norm also reduces QL-$L_{\infty}$'s query needs, and as a result, produces larger distortions than QL. We remark that removing the perturbation constraint will help in query reduction. Note that comparing to the query needs reported by \citet{chen2017zoo}, ZOO-$L_2$ reduces the query counts from more than 2 millions to 77,513, which is $1\sim2$ orders of magnitude.
	Compared with QL-$L_{\infty}$, which uses the same black-box optimization as Region Attack, we reduce the query needs by half. This indicates that only removing the distance loss in existing attacks does not produce efficient attacks;  {our Region Attack reduces the dimension of the attack space and allows fewer queries in gradient estimation.}
	
		
\begin{figure}[t!]
    \centering
    \includegraphics[width=\columnwidth]{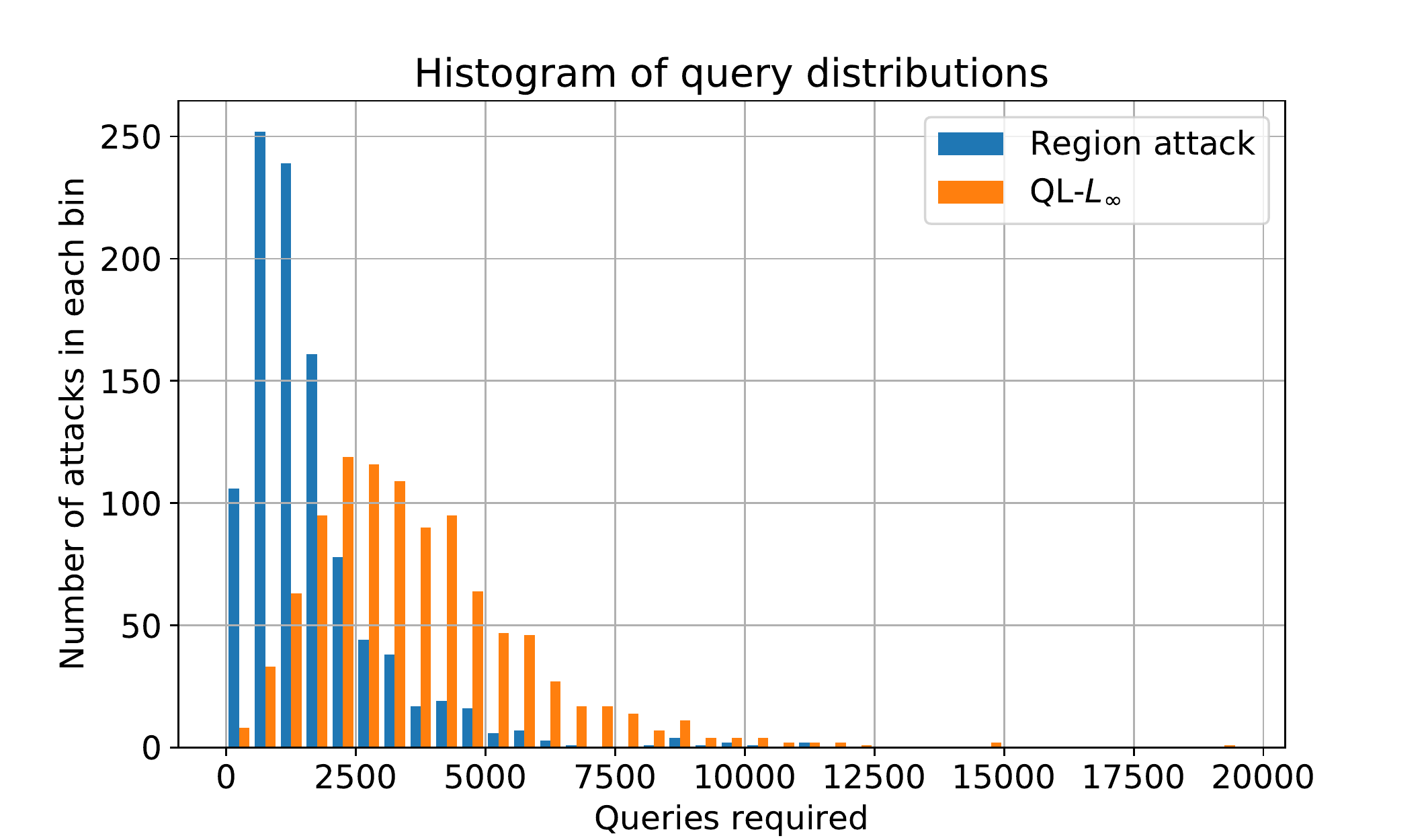}
    \caption{Query distributions of Region attack and QL-$L_{\infty} $ methods on 1000 targeted attacks with $\eta=0.7, h=64$. The queries are divided into 39 bins with each bin spanning a width of 500 queries. Blue bar shows the query distribution for Region Attack; Orange bar shows queries for QL-$L_{\infty}$.
    }\label{fig2:hist}
\end{figure}

\subsubsection{Query distribution comparison}
Since QL-$L_{\infty}$ is the strongest baseline method, we compare its query distributions with Region Attack in Fig. \ref{fig2:hist}. 
	Note that the maximal and minimal number of queries for QL-$L_{\infty}$ and Region Attack are 19431 and 51 respectively; we thus divide queries of 1000 attacks into 39 bins with each bin spanning a width of 500 queries. The highest blue bar shows that 252 targeted attacks are completed within queries $[500, 1000)$. 
	Dividing the histogram by query counts of 2500, 83.6\% attacks of Region Attack are less than 2500 while it is only 31.8\%  for QL-$L_{\infty}$.

\begin{figure}[t!]
\centering
    \includegraphics[width=\columnwidth]{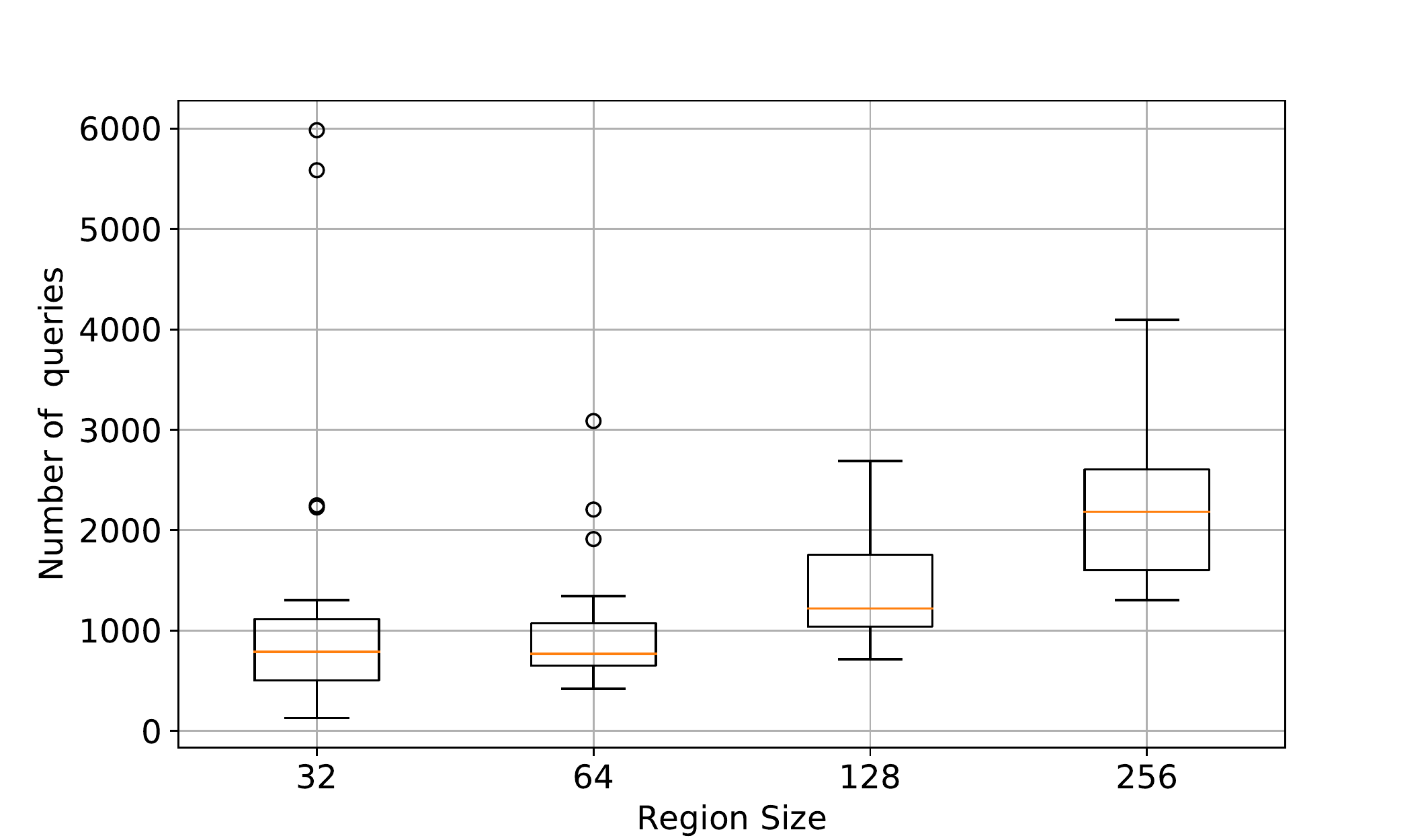}
    \caption{  Query comparison of different region sizes under 20 targeted attacks. Centerline represents the median. Box extents show 25th and 75th percentile. Whisker extents illustrate maximum and minimum values. Black circles show outliers that are too far from the median such as 6000 queries for target class \emph{tench} under size 32.
    }\label{fig3:size}
\end{figure}

\begin{table}[t!b]
    \centering
    \caption{Comparisons between different region sizes on success rate, average queries against InceptionV3  on 20 targeted attacks with $\eta=0.4$.}
    \label{table:2-size}
    \begin{tabular}{l|c|c|c}
        \hline
        & success rate & avg. queries & size selection cost\\ \hline
        $(w,h)$=32 & 100\% & 1344.00 & -\\ \hline
        $(w,h)$=64& 100\% & 1012.20 & -\\ \hline
        $(w,h)$=128& 100\% &  1408.05& -\\ \hline
        $(w,h)$=256 &  {100\% }&2297.40 & -\\ \hline
        size selection&  100\%& 1033.25 & 200 \\ \hline
    \end{tabular}
\end{table}
	
\subsubsection{Influence of size selection on query complexity}
We compare the active size selection algorithm with fixed sizes chosen from \{32, 64, 128, 256\} and report query complexity on attacking the first 20 classes of ImageNet which are \emph{ tench, goldfish, great white shark, tiger shark, hammerhead shark, electric ray, stingray, cock, hen, ostrich, brambling, goldfinch, house finch, junco, indigo bird, robin, bulbul, jay, magpie, chickadee}. 
	
	The average results are shown in  Table \ref{table:2-size}. The active selection costs are set to be 200 queries for each targeted attack. In Table \ref{table:2-size}, we find that the active selection strategy leads to lower average query counts than using a fixed size selected from \{32,128,256\}, but marginally higher average query counts than using size 64.  This observation is consistent with Table \ref{table:1} in which the active selection strategy reports lower average query needs than attack with a fixed size $w, h=32$.
	This indicates that when one does not have prior knowledge about the suitable size for an attack, our size selection strategy will provide a satisfactory choice.
	
	Besides, the average number of queries required for region size 64 is 1012, 2 times faster than region size 256. This observation validates our opinion that the target model is robust to the size of the object and a smaller but suitable region will lead to query reduction. Moreover,  Fig. \ref{fig3:size} shows the detailed property of comparisons among different sizes. Observing from Fig. \ref{fig3:size}, we remark that excessively small region size such as $w=h=32$ will lead to unstable results.
\begin{figure}[t!]
    \centering
    \subfloat[Average query for  different momentum ]{\includegraphics[width=\columnwidth]{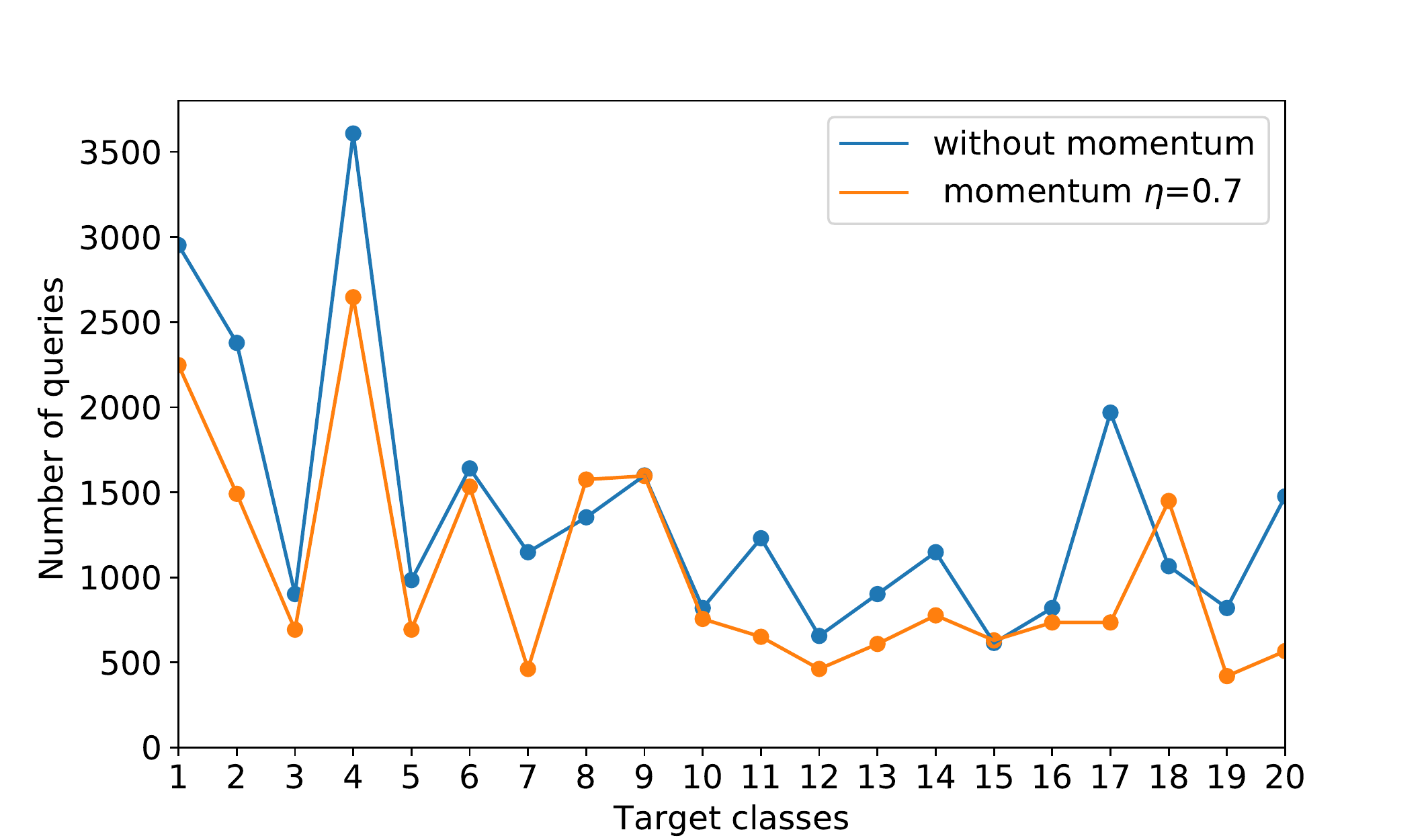}
        \label{fig4:momentum1}}
    \qquad
    \subfloat[Queries for twenty attacks]{\includegraphics[width=\columnwidth]{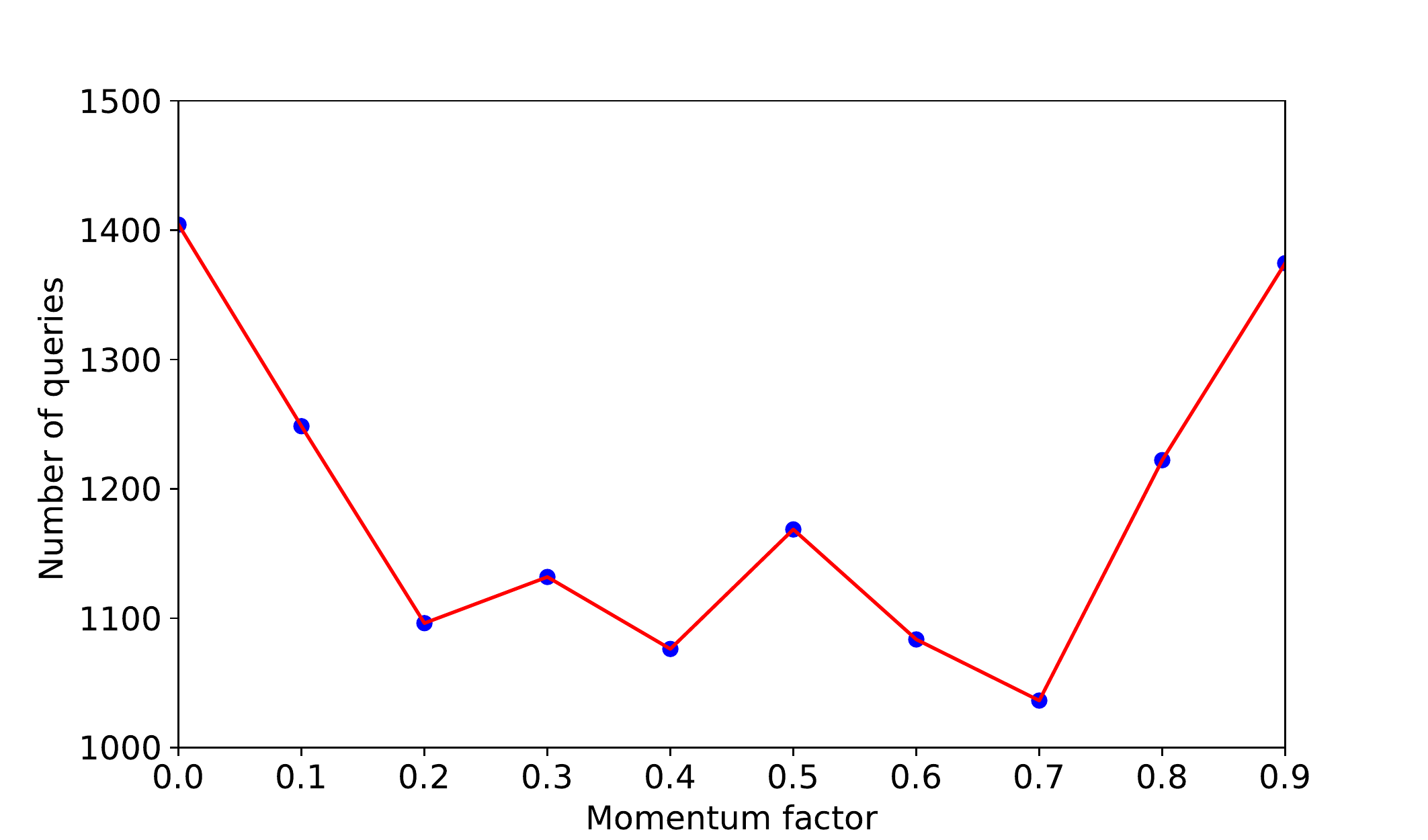}
        \label{fig4:momentum2}}
    \caption{Query complexity comparisons under different momentum factor. (a) Average queries for 20 targeted attacks under different momentum factor. (b) Comparisons of query for each attack with gradient ascent and momentum gradient ascent at $\eta=0.7$ .}
    \label{fig4:momentum}
\end{figure}

\subsubsection{Comparisons of SGD with and without Momentum}
	We implement attacks to test momentum method's influence on SGD in optimization.
	In Fig. \ref{fig4:momentum1}, we compare queries on 20 targeted attacks. The blue line shows the query needs for gradient ascent without momentum, which is above the yellow line showing query needs with the momentum factor $\eta=0.7$ most of the time. This observation corroborates that momentum method plays an important role in stabilizing the optimization thus leading to faster convergence. 
	Furthermore,  we examine different momentum factor $\eta$'s influence on query complexity. 
	For $\eta\in \{0, 0.1, ..., 0.9\}$, we report the average number of queries over 20 targeted attacks in Fig. \ref{fig4:momentum2}. One can find
	that the query count first goes down with the increase of momentum factor $\eta$ and then goes up when $\eta$ is larger than $0.7$. As we analyzed before, a large momentum factor will hinder the algorithm from exploring a new ascent direction, while a small momentum factor merely has the power to maintain the correct ascent direction.

\subsection{Results on Partial-information Attack}\label{sec:exp:partial-info}
In this battery of experiments, we evaluate the performance of Region Attack under partial-information settings. Since Region Attack and QL adopt different initialization methods, Region Attack uses random initialization and  QL uses examples from target class, we can not compare them directly. 
	
	We first evaluate Region Attack on InceptionV3.  Attacks under partial information are more difficult than the one under full information {since we have to carefully choose small learning rate for updating the adversarial image and mode $\sigma$ of search distribution to avoid stepping out of the TOP-K list.}  Assume we have access to the TOP-20 predictions, the fifth predictions of InceptionV3 on the gray image is \emph{whistle} with confidence 1.21\%. 
	We set momentum factor $\eta=0.1$, region size $h=64$ and batch size $n =40$ and maximize its score until it is placed on the 1st place. 
	In the experiments, we find that Region Attack under the Gaussian search distribution converges slowly with vibrations. This implies that adding small noises to pixels does not effectively help search the useful gradient direction. We replace it with the Laplace distribution which is sharper compared to Gaussian and observe better performance. Within $10820$ queries, label \emph{whistle} is ranked on the first place. The Laplace distribution will be used in the followed online experiments.
\begin{figure}[t!]
    \centering
    \subfloat[Initial results]{\includegraphics[width=\columnwidth]    {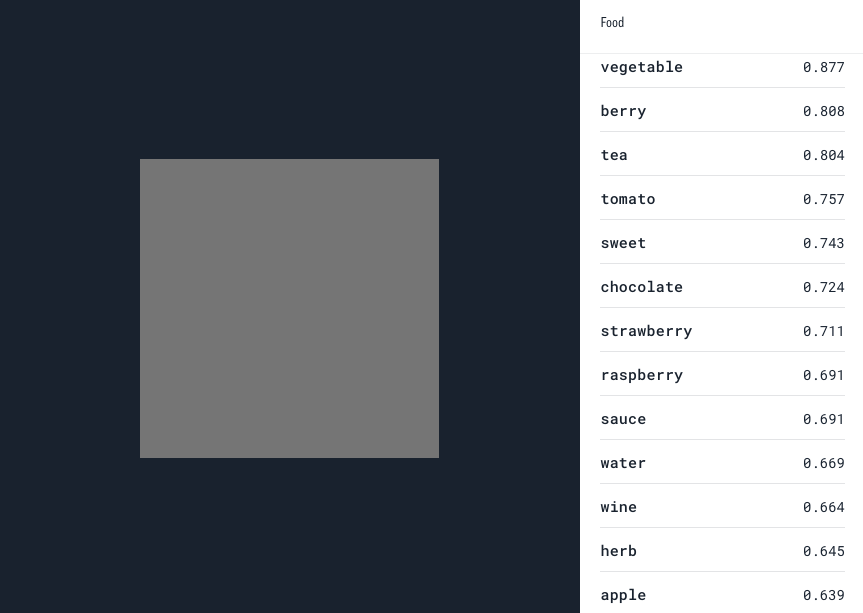} 
        \label{fig6:clarifai-food1}}
    \qquad
    \subfloat[Targeted attack on \emph{apple}]{\includegraphics[width=\columnwidth]{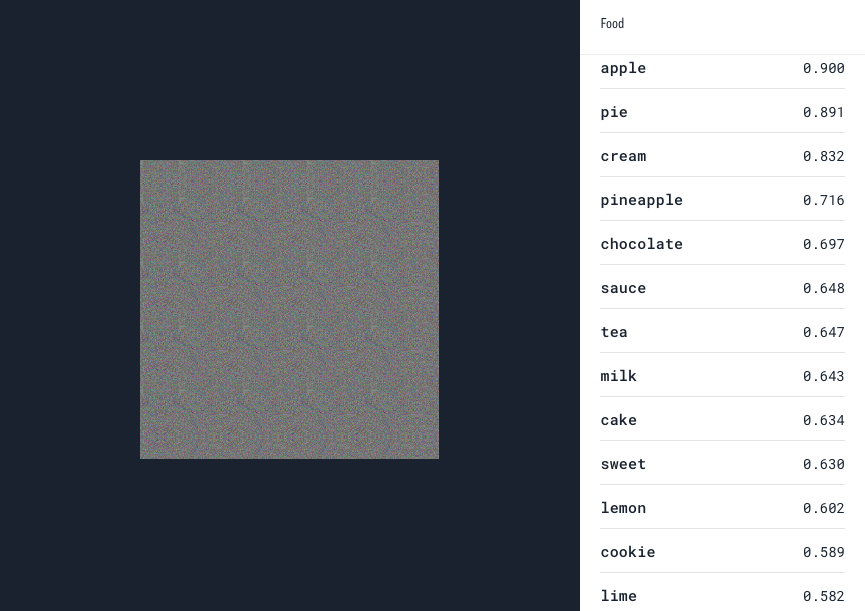}
        \label{fig6:clarifai-food2}}
    \caption{The Clarifai food detection API's prediction results on an adversarial image produced by Region Attack with $\eta=0.4$ and query need as 5,000.}\label{fig6:clarifai-food}
\end{figure}
\begin{figure}[p]
		\centering
		\subfloat[Initial results]{\includegraphics[width=0.8\columnwidth]{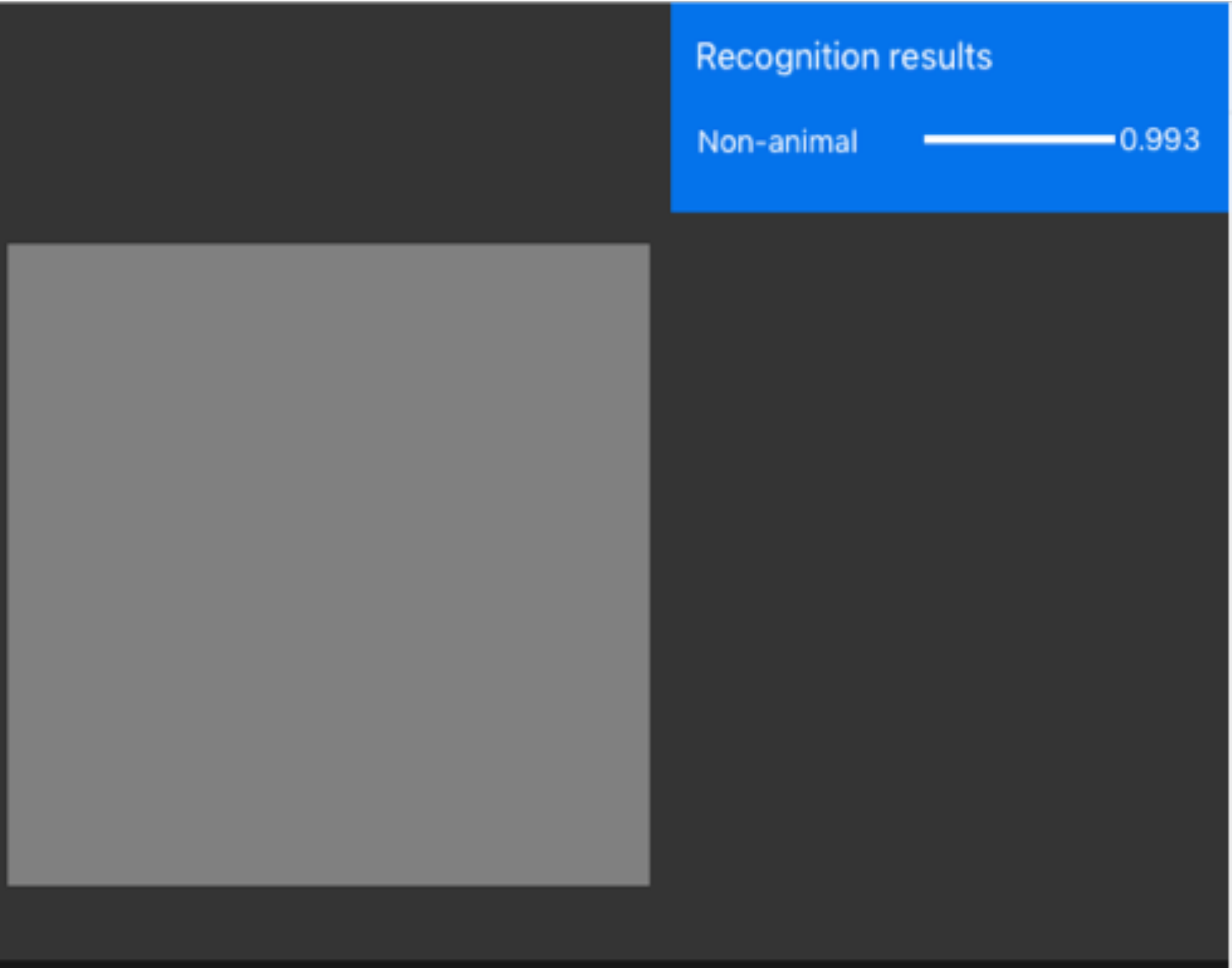}
			\label{fig7:baidu-animal-1}}
		\\
		\subfloat[Intermediate adversarial example]{\includegraphics[width=0.8\columnwidth]{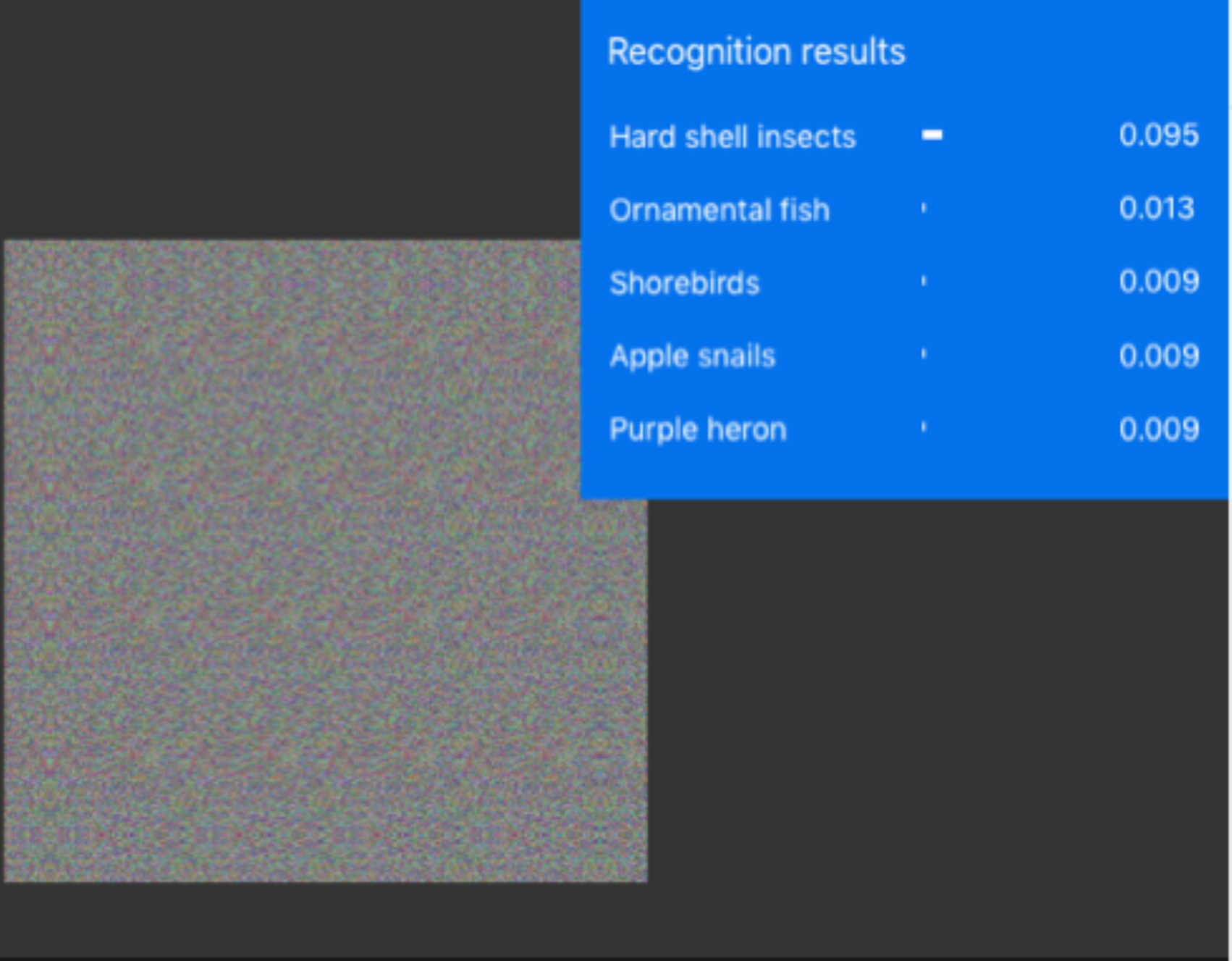}
			\label{fig7:baidu-animal-2}}
		\\
		\subfloat[Targeted attack on \emph{Purple heron}]{\includegraphics[width=0.8\columnwidth]{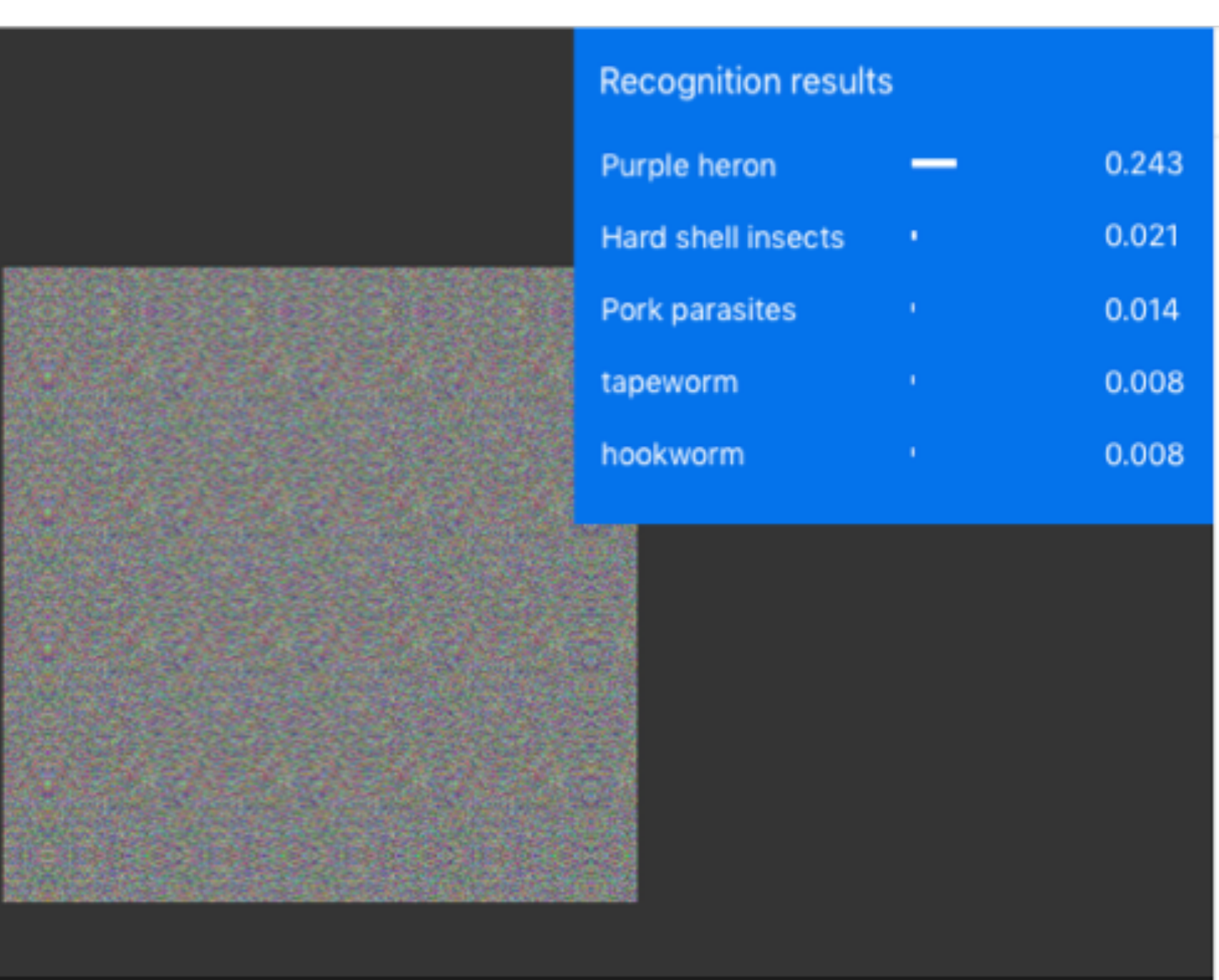}
			\label{fig7:baidu-animal-3}}
		\caption{Region Attack on Baidu AI Platform with $\eta$=0.4. {(a):} original prediction on gray image; {(b):} the intermediate adversarial example when the label \emph{non-animal} is perturbed out of TOP-5 predictions with 1500 queries; {(c):} extra 2400 queries are needed to raise target label \emph{Purple heron} to TOP-1 prediction with score $0.0421$. With further 3,000 queries, the score for \emph{Purple heron} is increased to $0.243$, which is ten-folds larger than that of second label.}
		\label{fig7:baidu-animal}
	\end{figure}
    
\subsection{Real-world Attack on Clarifai and Baidu Cloud APIs}
	We test Region Attack against real-world image recognition systems in this section. We choose food detection API from Clarifai which offers label detection service to public, and animal identification API from Baidu AI platform which provides fine-grained object identification service as target models.
	
	\subsubsection{Results on Clarifai}
	Under partial-information settings, attacking Clarifai is more challenging than attacking InceptionV3. On one hand, the total number of labels is large and unknown. On the other hand, Clarifai provides multi-label prediction results, and the scores of returned list of labels are not summed to be 1. Furthermore, queries on Clarifai are expensive  US\$1.2 for 1,000 queries. We show that Region Attack reports promising results even under this hard scenario.
	
	We implement the partial-information attack discussed in Section \ref{sec:partial-info}. Since we cannot access to  any natural input example and legal label, we select the gray image as the initialization of adversarial examples and obtain the initial results shown in Fig. \ref{fig6:clarifai-food1}. In Fig. \ref{fig6:clarifai-food1}, original image are labeled as \emph{vegetable} by Clarifai with confidence 87.7\%.  We choose 13th label \emph{apple} as target label to perform targeted attack. With around 5,000 queries to the food API, we obtain the adversarial image that is labeled as \emph{apple} with 90\% confidence as shown in Fig. \ref{fig6:clarifai-food2}.

\subsubsection{Results on Baidu AI platform}
	We test the animal identification API on Baidu Cloud and report the results in Fig. \ref{fig7:baidu-animal}. Fig. \ref{fig7:baidu-animal-1}  gives the original prediction of the gray image, which is \emph{non-animal}. We treat it as a binary classification problem by first minimizing the score of \emph{non-animal} until it falls out of TOP-5 predictions. Fig. \ref{fig7:baidu-animal-2} shows identification result when \emph{non-animal} falls out of TOP-5.  We pick \emph{Purple heron} as our target label which has the lowest probability among TOP-5 predictions and maximize its score until its score is raised to TOP-1. 
	Fig. \ref{fig7:baidu-animal-3}  gives predictions when adversarial example are classified into \emph{Purple heron} with high probability. 
		
\section{Conclusion}\label{sec:conclusion}
{In this work, we introduce a new attack approach called \emph{input-free} attack. Under this setting, we propose \emph{Region Attack}, a black-box attack algorithm that can significantly reduce the query complexity. 
Our extensive experiments show that our algorithm is quite effective: it can achieve a 100\% targeted-attack success rate with less than 2,000 queries on average on ImageNet dataset. Besides, it can also defeat some real-world commercial classification systems.  
Our results clearly suggest that, compared with traditional adversarial attacks that require the adversarial example to be visually similar to the original example, the \textit{input-free} attack is more effective and practical in query-limited settings.

\begin{acks}
The authors would like to thank Sam Bretheim and the anonymous reviewers for their valuable feedback and helpful suggestions.
The work is supported by Australian Research Council Projects FL-170100117 and DP-180103424, National Natural Science Foundation of China (61772508, U1713213), Guangdong Technology Project (2016B010108010, 2016B010125003, 2017B010110007), CAS Key Technology Talent Program, Shenzhen Engineering Laboratory for 3D Content Generating Technologies (NO. [2017]476).
\end{acks}

\bibliographystyle{ACM-Reference-Format}
\balance


\begin{figure*}[]
	\centering
	\includegraphics[width=2\columnwidth]{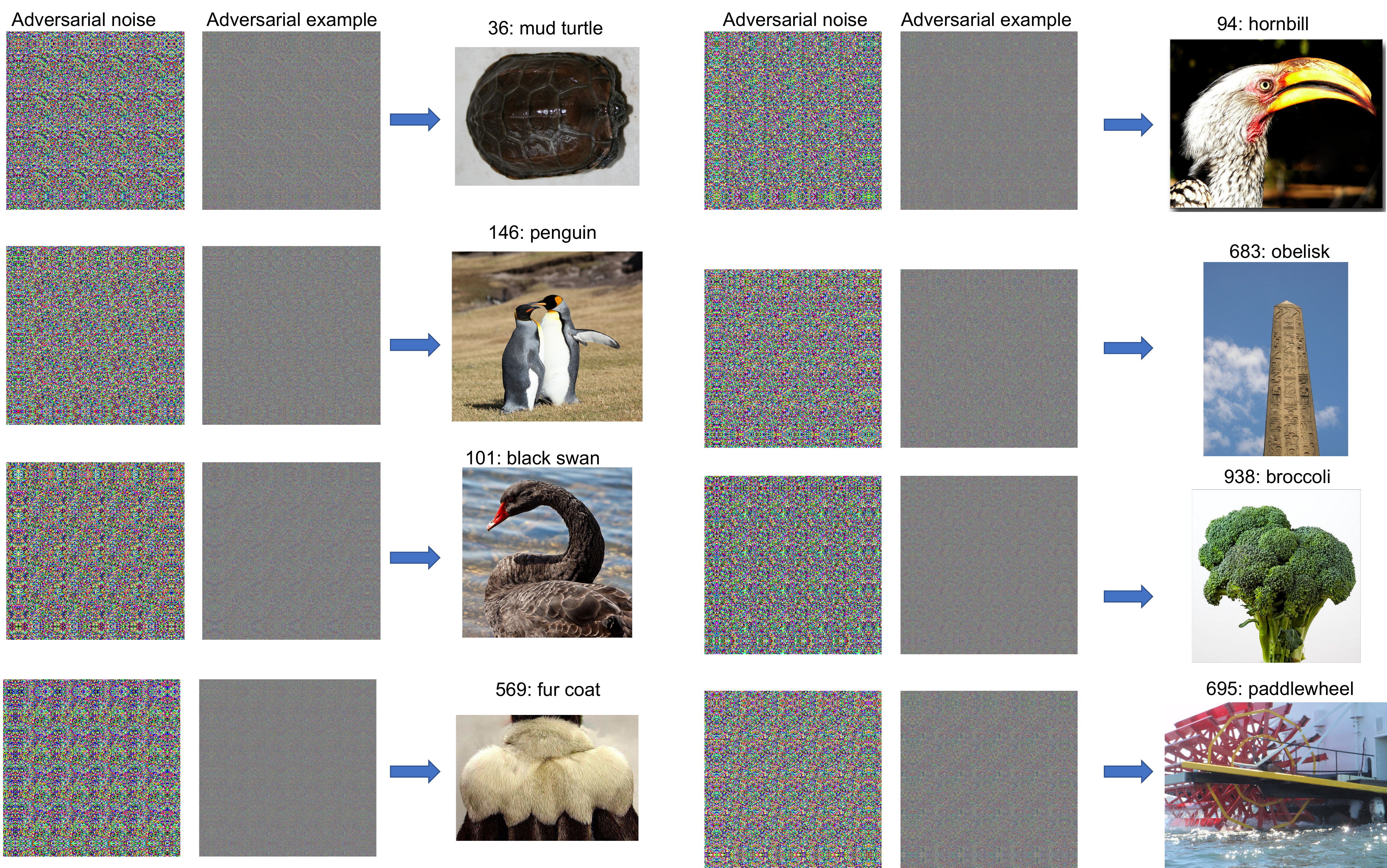}
	\caption{Adversarial noises, examples for InceptionV3 and ground-truth examples from ImageNet. Adversarial examples are obtained by putting the gray image and adversarial noises together. The number before a label indicates the index of the label in ImageNet dataset. 
	}\label{fig:app1:examples}
\end{figure*}

\newpage
\appendix
\section*{Appendix A. Adversarial examples}
Figure  \ref{fig:app1:examples} shows some adversarial examples for targeted attacks in ImageNet where clear texture patterns of target classes can be observed from the adversarial examples.


\begin{thebibliography}{25}


\ifx \showCODEN    \undefined \def \showCODEN     #1{\unskip}     \fi
\ifx \showDOI      \undefined \def \showDOI       #1{#1}\fi
\ifx \showISBNx    \undefined \def \showISBNx     #1{\unskip}     \fi
\ifx \showISBNxiii \undefined \def \showISBNxiii  #1{\unskip}     \fi
\ifx \showISSN     \undefined \def \showISSN      #1{\unskip}     \fi
\ifx \showLCCN     \undefined \def \showLCCN      #1{\unskip}     \fi
\ifx \shownote     \undefined \def \shownote      #1{#1}          \fi
\ifx \showarticletitle \undefined \def \showarticletitle #1{#1}   \fi
\ifx \showURL      \undefined \def \showURL       {\relax}        \fi
\providecommand\bibfield[2]{#2}
\providecommand\bibinfo[2]{#2}
\providecommand\natexlab[1]{#1}
\providecommand\showeprint[2][]{arXiv:#2}

\bibitem[\protect\citeauthoryear{Abadi, Barham, Chen, Chen, Davis, Dean, Devin,
  Ghemawat, Irving, Isard, et~al\mbox{.}}{Abadi et~al\mbox{.}}{2016}]%
        {abadi2016tensorflow}
\bibfield{author}{\bibinfo{person}{Mart{\'\i}n Abadi}, \bibinfo{person}{Paul
  Barham}, \bibinfo{person}{Jianmin Chen}, \bibinfo{person}{Zhifeng Chen},
  \bibinfo{person}{Andy Davis}, \bibinfo{person}{Jeffrey Dean},
  \bibinfo{person}{Matthieu Devin}, \bibinfo{person}{Sanjay Ghemawat},
  \bibinfo{person}{Geoffrey Irving}, \bibinfo{person}{Michael Isard},
  {et~al\mbox{.}}} \bibinfo{year}{2016}\natexlab{}.
\newblock \showarticletitle{TensorFlow: a system for large-scale machine
  learning}. In \bibinfo{booktitle}{{\em Proceedings of the 12th USENIX
  conference on Operating Systems Design and Implementation}}. USENIX
  Association, \bibinfo{pages}{265--283}.
\newblock


\bibitem[\protect\citeauthoryear{Barreno, Nelson, Joseph, and Tygar}{Barreno
  et~al\mbox{.}}{2010}]%
        {barreno2010security}
\bibfield{author}{\bibinfo{person}{Marco Barreno}, \bibinfo{person}{Blaine
  Nelson}, \bibinfo{person}{Anthony~D Joseph}, {and} \bibinfo{person}{JD
  Tygar}.} \bibinfo{year}{2010}\natexlab{}.
\newblock \showarticletitle{The security of machine learning}.
\newblock \bibinfo{journal}{{\em Machine Learning\/}} \bibinfo{volume}{81},
  \bibinfo{number}{2} (\bibinfo{year}{2010}), \bibinfo{pages}{121--148}.
\newblock


\bibitem[\protect\citeauthoryear{Barreno, Nelson, Sears, Joseph, and
  Tygar}{Barreno et~al\mbox{.}}{2006}]%
        {barreno2006can}
\bibfield{author}{\bibinfo{person}{Marco Barreno}, \bibinfo{person}{Blaine
  Nelson}, \bibinfo{person}{Russell Sears}, \bibinfo{person}{Anthony~D Joseph},
  {and} \bibinfo{person}{J~Doug Tygar}.} \bibinfo{year}{2006}\natexlab{}.
\newblock \showarticletitle{Can machine learning be secure?}. In
  \bibinfo{booktitle}{{\em Proceedings of the 2006 ACM Symposium on
  Information, computer and communications security}}. ACM,
  \bibinfo{pages}{16--25}.
\newblock


\bibitem[\protect\citeauthoryear{Brendel, Rauber, and Bethge}{Brendel
  et~al\mbox{.}}{2017}]%
        {brendel2017decision}
\bibfield{author}{\bibinfo{person}{Wieland Brendel}, \bibinfo{person}{Jonas
  Rauber}, {and} \bibinfo{person}{Matthias Bethge}.}
  \bibinfo{year}{2017}\natexlab{}.
\newblock \showarticletitle{Decision-Based Adversarial Attacks: Reliable
  Attacks Against Black-Box Machine Learning Models}.
\newblock \bibinfo{journal}{{\em arXiv preprint arXiv:1712.04248\/}}
  (\bibinfo{year}{2017}).
\newblock


\bibitem[\protect\citeauthoryear{Carlini and Wagner}{Carlini and
  Wagner}{2017}]%
        {carlini2017towards}
\bibfield{author}{\bibinfo{person}{Nicholas Carlini} {and}
  \bibinfo{person}{David Wagner}.} \bibinfo{year}{2017}\natexlab{}.
\newblock \showarticletitle{Towards evaluating the robustness of neural
  networks}. In \bibinfo{booktitle}{{\em IEEE Symposium on Security and Privacy
  (SP)}}. IEEE, \bibinfo{pages}{39--57}.
\newblock


\bibitem[\protect\citeauthoryear{Chen, Zhang, Sharma, Yi, and Hsieh}{Chen
  et~al\mbox{.}}{2017}]%
        {chen2017zoo}
\bibfield{author}{\bibinfo{person}{Pin-Yu Chen}, \bibinfo{person}{Huan Zhang},
  \bibinfo{person}{Yash Sharma}, \bibinfo{person}{Jinfeng Yi}, {and}
  \bibinfo{person}{Cho-Jui Hsieh}.} \bibinfo{year}{2017}\natexlab{}.
\newblock \showarticletitle{Zoo: Zeroth order optimization based black-box
  attacks to deep neural networks without training substitute models}. In
  \bibinfo{booktitle}{{\em Proceedings of the 10th ACM Workshop on Artificial
  Intelligence and Security}}. ACM, \bibinfo{pages}{15--26}.
\newblock


\bibitem[\protect\citeauthoryear{Deng, Dong, Socher, Li, Li, and Fei-Fei}{Deng
  et~al\mbox{.}}{2009}]%
        {deng2009imagenet}
\bibfield{author}{\bibinfo{person}{Jia Deng}, \bibinfo{person}{Wei Dong},
  \bibinfo{person}{Richard Socher}, \bibinfo{person}{Li-Jia Li},
  \bibinfo{person}{Kai Li}, {and} \bibinfo{person}{Li Fei-Fei}.}
  \bibinfo{year}{2009}\natexlab{}.
\newblock \showarticletitle{Imagenet: A large-scale hierarchical image
  database}. In \bibinfo{booktitle}{{\em IEEE Conference on Computer Vision and
  Pattern Recognition}}. IEEE, \bibinfo{pages}{248--255}.
\newblock


\bibitem[\protect\citeauthoryear{Dong, Liao, Pang, Su, Zhu, Hu, and Li}{Dong
  et~al\mbox{.}}{2018}]%
        {dong2018boosting}
\bibfield{author}{\bibinfo{person}{Yinpeng Dong}, \bibinfo{person}{Fangzhou
  Liao}, \bibinfo{person}{Tianyu Pang}, \bibinfo{person}{Hang Su},
  \bibinfo{person}{Jun Zhu}, \bibinfo{person}{Xiaolin Hu}, {and}
  \bibinfo{person}{Jianguo Li}.} \bibinfo{year}{2018}\natexlab{}.
\newblock \showarticletitle{Boosting adversarial attacks with momentum}.
\newblock \bibinfo{journal}{{\em arXiv preprint\/}} (\bibinfo{year}{2018}).
\newblock


\bibitem[\protect\citeauthoryear{Engstrom, Tsipras, Schmidt, and
  Madry}{Engstrom et~al\mbox{.}}{2017}]%
        {engstrom2017rotation}
\bibfield{author}{\bibinfo{person}{Logan Engstrom}, \bibinfo{person}{Dimitris
  Tsipras}, \bibinfo{person}{Ludwig Schmidt}, {and} \bibinfo{person}{Aleksander
  Madry}.} \bibinfo{year}{2017}\natexlab{}.
\newblock \showarticletitle{A Rotation and a Translation Suffice: Fooling CNNs
  with Simple Transformations}.
\newblock \bibinfo{journal}{{\em arXiv preprint arXiv:1712.02779\/}}
  (\bibinfo{year}{2017}).
\newblock


\bibitem[\protect\citeauthoryear{Goodfellow, Shlens, and Szegedy}{Goodfellow
  et~al\mbox{.}}{2014}]%
        {goodfellow2014explaining}
\bibfield{author}{\bibinfo{person}{Ian~J Goodfellow}, \bibinfo{person}{Jonathon
  Shlens}, {and} \bibinfo{person}{Christian Szegedy}.}
  \bibinfo{year}{2014}\natexlab{}.
\newblock \showarticletitle{Explaining and harnessing adversarial examples}.
\newblock \bibinfo{journal}{{\em arXiv preprint arXiv:1412.6572\/}}
  (\bibinfo{year}{2014}).
\newblock


\bibitem[\protect\citeauthoryear{Hayes and Danezis}{Hayes and Danezis}{2017}]%
        {hayes2017machine}
\bibfield{author}{\bibinfo{person}{Jamie Hayes} {and} \bibinfo{person}{George
  Danezis}.} \bibinfo{year}{2017}\natexlab{}.
\newblock \showarticletitle{Machine learning as an adversarial service:
  Learning black-box adversarial examples}.
\newblock \bibinfo{journal}{{\em arXiv preprint arXiv:1708.05207\/}}
  (\bibinfo{year}{2017}).
\newblock


\bibitem[\protect\citeauthoryear{Ilyas, Engstrom, Athalye, and Lin}{Ilyas
  et~al\mbox{.}}{2018}]%
        {ilyas2018black}
\bibfield{author}{\bibinfo{person}{Andrew Ilyas}, \bibinfo{person}{Logan
  Engstrom}, \bibinfo{person}{Anish Athalye}, {and} \bibinfo{person}{Jessy
  Lin}.} \bibinfo{year}{2018}\natexlab{}.
\newblock \showarticletitle{Black-box Adversarial Attacks with Limited Queries
  and Information}.
\newblock \bibinfo{journal}{{\em arXiv preprint arXiv:1804.08598\/}}
  (\bibinfo{year}{2018}).
\newblock


\bibitem[\protect\citeauthoryear{Liu, Chen, Liu, and Song}{Liu
  et~al\mbox{.}}{2016}]%
        {liu2016delving}
\bibfield{author}{\bibinfo{person}{Yanpei Liu}, \bibinfo{person}{Xinyun Chen},
  \bibinfo{person}{Chang Liu}, {and} \bibinfo{person}{Dawn Song}.}
  \bibinfo{year}{2016}\natexlab{}.
\newblock \showarticletitle{Delving into transferable adversarial examples and
  black-box attacks}.
\newblock \bibinfo{journal}{{\em arXiv preprint arXiv:1611.02770\/}}
  (\bibinfo{year}{2016}).
\newblock


\bibitem[\protect\citeauthoryear{Madry, Makelov, Schmidt, Tsipras, and
  Vladu}{Madry et~al\mbox{.}}{2017}]%
        {madry2017towards}
\bibfield{author}{\bibinfo{person}{Aleksander Madry},
  \bibinfo{person}{Aleksandar Makelov}, \bibinfo{person}{Ludwig Schmidt},
  \bibinfo{person}{Dimitris Tsipras}, {and} \bibinfo{person}{Adrian Vladu}.}
  \bibinfo{year}{2017}\natexlab{}.
\newblock \showarticletitle{Towards deep learning models resistant to
  adversarial attacks}.
\newblock \bibinfo{journal}{{\em arXiv preprint arXiv:1706.06083\/}}
  (\bibinfo{year}{2017}).
\newblock


\bibitem[\protect\citeauthoryear{Moosavi~Dezfooli, Fawzi, and
  Frossard}{Moosavi~Dezfooli et~al\mbox{.}}{2016}]%
        {moosavi2016deepfool}
\bibfield{author}{\bibinfo{person}{Seyed~Mohsen Moosavi~Dezfooli},
  \bibinfo{person}{Alhussein Fawzi}, {and} \bibinfo{person}{Pascal Frossard}.}
  \bibinfo{year}{2016}\natexlab{}.
\newblock \showarticletitle{Deepfool: a simple and accurate method to fool deep
  neural networks}. In \bibinfo{booktitle}{{\em Proceedings of 2016 IEEE
  Conference on Computer Vision and Pattern Recognition (CVPR)}}.
\newblock


\bibitem[\protect\citeauthoryear{Nguyen, Yosinski, and Clune}{Nguyen
  et~al\mbox{.}}{2015}]%
        {nguyen2015deep}
\bibfield{author}{\bibinfo{person}{Anh Nguyen}, \bibinfo{person}{Jason
  Yosinski}, {and} \bibinfo{person}{Jeff Clune}.}
  \bibinfo{year}{2015}\natexlab{}.
\newblock \showarticletitle{Deep neural networks are easily fooled: High
  confidence predictions for unrecognizable images}. In
  \bibinfo{booktitle}{{\em Proceedings of the IEEE Conference on Computer
  Vision and Pattern Recognition}}. \bibinfo{pages}{427--436}.
\newblock


\bibitem[\protect\citeauthoryear{Papernot, McDaniel, Goodfellow, Jha, Celik,
  and Swami}{Papernot et~al\mbox{.}}{2017}]%
        {papernot2017practical}
\bibfield{author}{\bibinfo{person}{Nicolas Papernot}, \bibinfo{person}{Patrick
  McDaniel}, \bibinfo{person}{Ian Goodfellow}, \bibinfo{person}{Somesh Jha},
  \bibinfo{person}{Z~Berkay Celik}, {and} \bibinfo{person}{Ananthram Swami}.}
  \bibinfo{year}{2017}\natexlab{}.
\newblock \showarticletitle{Practical black-box attacks against machine
  learning}. In \bibinfo{booktitle}{{\em Proceedings of the 2017 ACM on Asia
  Conference on Computer and Communications Security}}. ACM,
  \bibinfo{pages}{506--519}.
\newblock


\bibitem[\protect\citeauthoryear{Papernot, McDaniel, Jha, Fredrikson, Celik,
  and Swami}{Papernot et~al\mbox{.}}{2016}]%
        {papernot2016limitations}
\bibfield{author}{\bibinfo{person}{Nicolas Papernot}, \bibinfo{person}{Patrick
  McDaniel}, \bibinfo{person}{Somesh Jha}, \bibinfo{person}{Matt Fredrikson},
  \bibinfo{person}{Z~Berkay Celik}, {and} \bibinfo{person}{Ananthram Swami}.}
  \bibinfo{year}{2016}\natexlab{}.
\newblock \showarticletitle{The limitations of deep learning in adversarial
  settings}. In \bibinfo{booktitle}{{\em IEEE European Symposium on Security
  and Privacy (EuroS\&P)}}. IEEE, \bibinfo{pages}{372--387}.
\newblock


\bibitem[\protect\citeauthoryear{Russakovsky, Deng, Su, Krause, Satheesh, Ma,
  Huang, Karpathy, Khosla, Bernstein, et~al\mbox{.}}{Russakovsky
  et~al\mbox{.}}{2015}]%
        {russakovsky2015imagenet}
\bibfield{author}{\bibinfo{person}{Olga Russakovsky}, \bibinfo{person}{Jia
  Deng}, \bibinfo{person}{Hao Su}, \bibinfo{person}{Jonathan Krause},
  \bibinfo{person}{Sanjeev Satheesh}, \bibinfo{person}{Sean Ma},
  \bibinfo{person}{Zhiheng Huang}, \bibinfo{person}{Andrej Karpathy},
  \bibinfo{person}{Aditya Khosla}, \bibinfo{person}{Michael Bernstein},
  {et~al\mbox{.}}} \bibinfo{year}{2015}\natexlab{}.
\newblock \showarticletitle{Imagenet large scale visual recognition challenge}.
\newblock \bibinfo{journal}{{\em International Journal of Computer Vision\/}}
  \bibinfo{volume}{115}, \bibinfo{number}{3} (\bibinfo{year}{2015}),
  \bibinfo{pages}{211--252}.
\newblock


\bibitem[\protect\citeauthoryear{Salimans, Ho, Chen, Sidor, and
  Sutskever}{Salimans et~al\mbox{.}}{2017}]%
        {salimans2017evolution}
\bibfield{author}{\bibinfo{person}{Tim Salimans}, \bibinfo{person}{Jonathan
  Ho}, \bibinfo{person}{Xi Chen}, \bibinfo{person}{Szymon Sidor}, {and}
  \bibinfo{person}{Ilya Sutskever}.} \bibinfo{year}{2017}\natexlab{}.
\newblock \showarticletitle{Evolution strategies as a scalable alternative to
  reinforcement learning}.
\newblock \bibinfo{journal}{{\em arXiv preprint arXiv:1703.03864\/}}
  (\bibinfo{year}{2017}).
\newblock


\bibitem[\protect\citeauthoryear{Su, Vargas, and Kouichi}{Su
  et~al\mbox{.}}{2017}]%
        {su2017one}
\bibfield{author}{\bibinfo{person}{Jiawei Su},
  \bibinfo{person}{Danilo~Vasconcellos Vargas}, {and} \bibinfo{person}{Sakurai
  Kouichi}.} \bibinfo{year}{2017}\natexlab{}.
\newblock \showarticletitle{One pixel attack for fooling deep neural networks}.
\newblock \bibinfo{journal}{{\em arXiv preprint arXiv:1710.08864\/}}
  (\bibinfo{year}{2017}).
\newblock


\bibitem[\protect\citeauthoryear{Szegedy, Vanhoucke, Ioffe, Shlens, and
  Wojna}{Szegedy et~al\mbox{.}}{2016}]%
        {szegedy2016rethinking}
\bibfield{author}{\bibinfo{person}{Christian Szegedy}, \bibinfo{person}{Vincent
  Vanhoucke}, \bibinfo{person}{Sergey Ioffe}, \bibinfo{person}{Jon Shlens},
  {and} \bibinfo{person}{Zbigniew Wojna}.} \bibinfo{year}{2016}\natexlab{}.
\newblock \showarticletitle{Rethinking the inception architecture for computer
  vision}. In \bibinfo{booktitle}{{\em Proceedings of the IEEE Conference on
  Computer Vision and Pattern Recognition}}. \bibinfo{pages}{2818--2826}.
\newblock


\bibitem[\protect\citeauthoryear{Szegedy, Zaremba, Sutskever, Bruna, Erhan,
  Goodfellow, and Fergus}{Szegedy et~al\mbox{.}}{2014}]%
        {szegedy2014intriguing}
\bibfield{author}{\bibinfo{person}{Christian Szegedy},
  \bibinfo{person}{Wojciech Zaremba}, \bibinfo{person}{Ilya Sutskever},
  \bibinfo{person}{Joan Bruna}, \bibinfo{person}{Dumitru Erhan},
  \bibinfo{person}{Ian Goodfellow}, {and} \bibinfo{person}{Rob Fergus}.}
  \bibinfo{year}{2014}\natexlab{}.
\newblock \showarticletitle{Intriguing properties of neural networks}. In
  \bibinfo{booktitle}{{\em ICLR}}. Citeseer.
\newblock


\bibitem[\protect\citeauthoryear{Tu, Ting, Chen, Liu, Zhang, Yi, Hsieh, and
  Cheng}{Tu et~al\mbox{.}}{2018}]%
        {tu2018autozoom}
\bibfield{author}{\bibinfo{person}{Chun-Chen Tu}, \bibinfo{person}{Paishun
  Ting}, \bibinfo{person}{Pin-Yu Chen}, \bibinfo{person}{Sijia Liu},
  \bibinfo{person}{Huan Zhang}, \bibinfo{person}{Jinfeng Yi},
  \bibinfo{person}{Cho-Jui Hsieh}, {and} \bibinfo{person}{Shin-Ming Cheng}.}
  \bibinfo{year}{2018}\natexlab{}.
\newblock \showarticletitle{AutoZOOM: Autoencoder-based Zeroth Order
  Optimization Method for Attacking Black-box Neural Networks}.
\newblock \bibinfo{journal}{{\em arXiv preprint arXiv:1805.11770\/}}
  (\bibinfo{year}{2018}).
\newblock


\bibitem[\protect\citeauthoryear{Wierstra, Schaul, Glasmachers, Sun, Peters,
  and Schmidhuber}{Wierstra et~al\mbox{.}}{2014}]%
        {wierstra2014natural}
\bibfield{author}{\bibinfo{person}{Daan Wierstra}, \bibinfo{person}{Tom
  Schaul}, \bibinfo{person}{Tobias Glasmachers}, \bibinfo{person}{Yi Sun},
  \bibinfo{person}{Jan Peters}, {and} \bibinfo{person}{J{\"u}rgen
  Schmidhuber}.} \bibinfo{year}{2014}\natexlab{}.
\newblock \showarticletitle{Natural evolution strategies.}
\newblock \bibinfo{journal}{{\em Journal of Machine Learning Research\/}}
  \bibinfo{volume}{15}, \bibinfo{number}{1} (\bibinfo{year}{2014}),
  \bibinfo{pages}{949--980}.
\newblock


\end{thebibliography}

\end{document}